\documentclass{tADR2e}

\usepackage{natbib}
\usepackage{tabularx}
\usepackage{ltablex}

\usepackage{hyperref}

\begin{document}

\jvol{01} \jnum{01} \jyear{2020} \jmonth{January}

\articletype{Article}

\title{Robots Assembling Machines: Learning from the World Robot Summit 2018 Assembly Challenge}

\author{Felix von Drigalski$^{a}$$^{\ast}$\thanks{$^\ast$Corresponding author. Email: FvDrigalski@gmail.com \vspace{6pt}},
Christian Schlette$^{b}$, Martin Rudorfer$^{c}$, Nikolaus Correll$^{d}$,\\Joshua C. Triyonoputro$^{e}$,
Weiwei Wan$^{e}$, Tokuo Tsuji$^{f}$, and Tetsuyou Watanabe$^{f}$\\
\vspace{6pt}  
$^{a}${\em{OMRON SINIC X Corp., 5-24-5 Hongo, Bunkyo-ku, Tokyo 113-0033, Japan}};\\
$^{b}${\em{University of Southern Denmark, Campusvej 55, 5230 Odense, Denmark}};\\
$^{c}${\em{Technical University Berlin, PTZ 5, Pascalstr. 8-9, 10587 Berlin, Germany}};\\
$^{d}${\em{University of Colorado Boulder and Robotic Materials Inc., USA}};\\
$^{e}${\em{Osaka University, 1-3 Machikaneyamacho, Toyonaka, Osaka, Japan}};\\
$^{f}${\em{Kanazawa University, Kakuma, Kanazawa, Japan}};
\\
\vspace{6pt}\received{March 2019} }

\maketitle

\begin{abstract}
The Industrial Assembly Challenge at the World Robot Summit was held in 2018 to showcase the state-of-the-art of autonomous manufacturing systems.
The challenge included various tasks, such as bin picking, kitting, and assembly of standard industrial parts into 2D and 3D assemblies. 
Some of the tasks were only revealed at the competition itself, representing the challenge of ``level 5'' automation, i. e., programming and setting up an autonomous assembly system in less than one day. 
We conducted a survey among the teams that participated in the challenge and investigated aspects such as team composition, development costs, system setups as well as the teams' strategies and approaches.
An analysis of the survey results reveals that the competitors have been in two camps: those constructing conventional robotic work cells with off-the-shelf tools, and teams who mostly relied on custom-made end effectors and novel software approaches in combination with collaborative robots.
While both camps performed reasonably well, the winning team chose a middle ground in between, combining the efficiency of established play-back programming with the autonomy gained by CAD-based object detection and force control for assembly operations.

\medskip

\medskip

\end{abstract}


\section{Introduction}

Robotic competitions have a long history of being used as a benchmark for AI research~\cite{anderson2011robotics}. 
Posing a well-defined technical challenge provides an opportunity to coalesce the community around important problems, while providing concrete metrics that allow researchers to compare their results.
Notable examples include the Defense Advanced Research Projects Agency (DARPA) Grand Challenge~\cite{buehler20072005} and the DARPA Urban Challenge~\cite{buehler2009darpa}, which broadly established the research discipline of autonomous driving and had significant economical impact by leading to the creation of numerous start-ups and eventually every car manufacturer pursuing autonomy research.
While the DARPA challenges were government efforts, the challenge format is also used by industry to understand and promote the state of the art. 
Here, the Amazon Robotics Challenge~\cite{correll2018analysis} has seen multiple iterations pushing the boundary of grasping retail items from tight packings. 
The US government has also shown interest in manipulation, albeit with a focus on manufacturing, with the National Institute of Standards and Technology (NIST) contributing to the First\footnote{\url{http://www.rhgm.org/activities/competition_iros2017/}, accessed 2019-07-02.} and Second Robotic Manipulation Challenges~\cite{sun2018robotic}, parts of which have been a precursor to the World Robot Summit~2018 Assembly Challenge, which is the subject of this paper. 

\begin{figure}[!htb]
    \centering
    \includegraphics[height=1.35in]{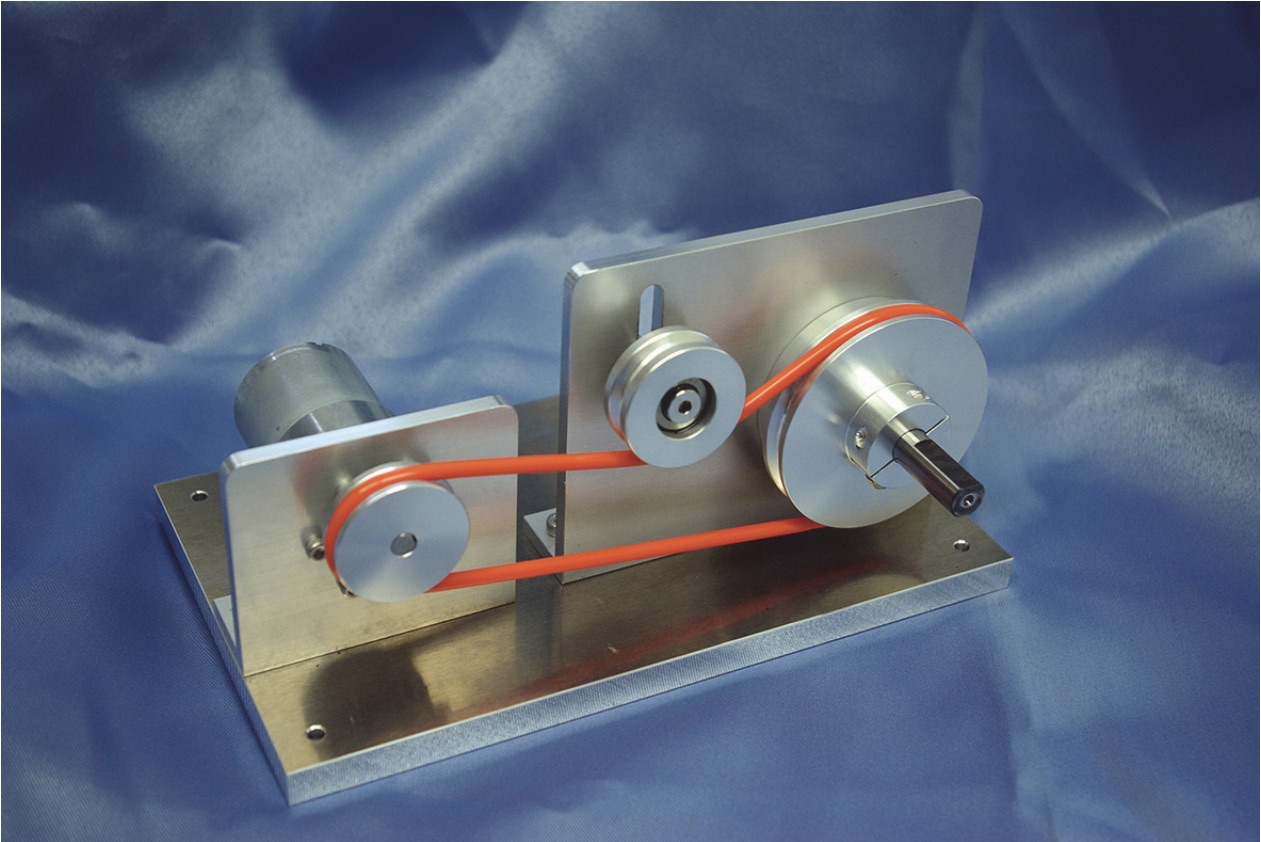}
    \includegraphics[height=1.35in,clip]{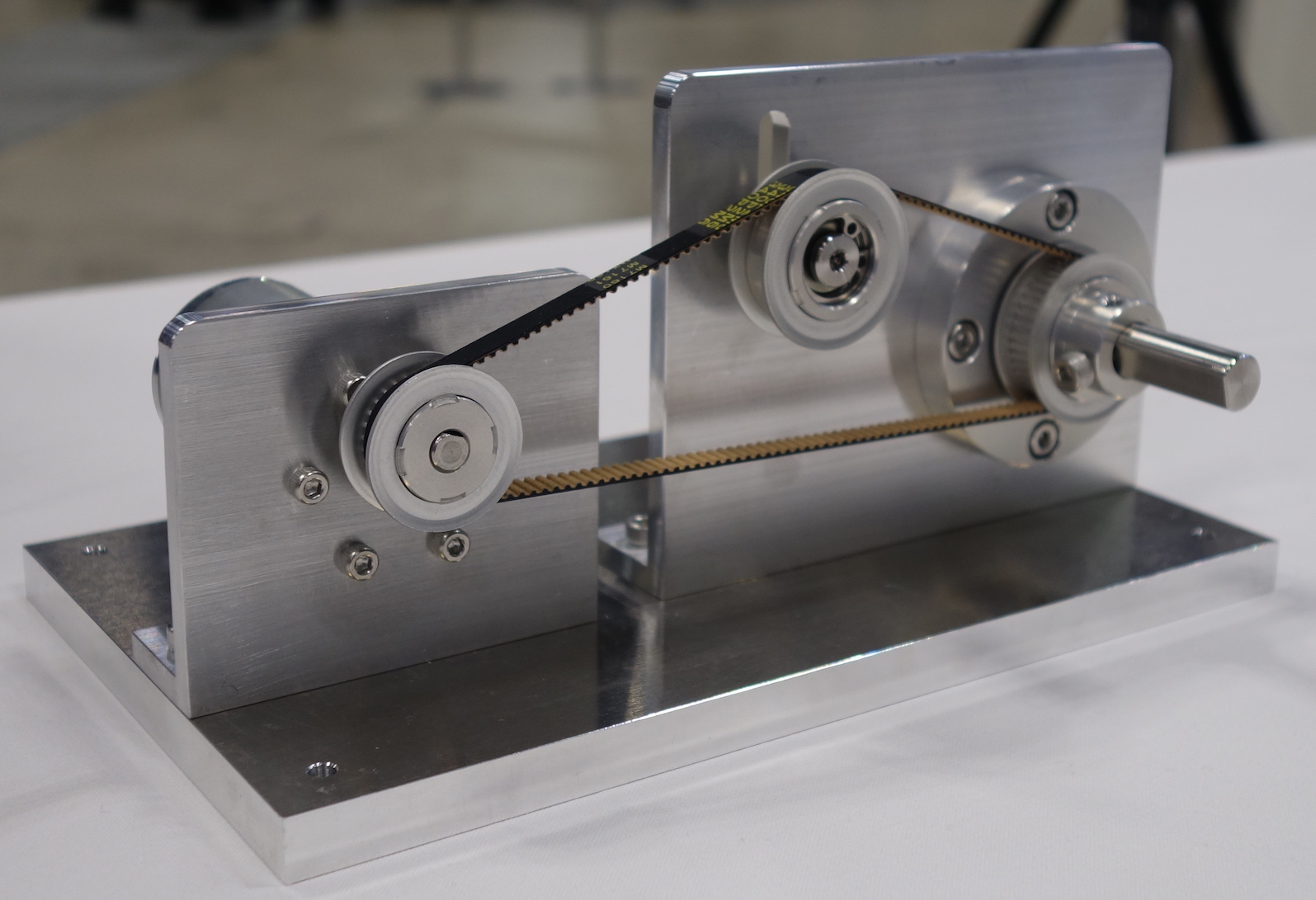}
    \includegraphics[height=1.35in,clip]{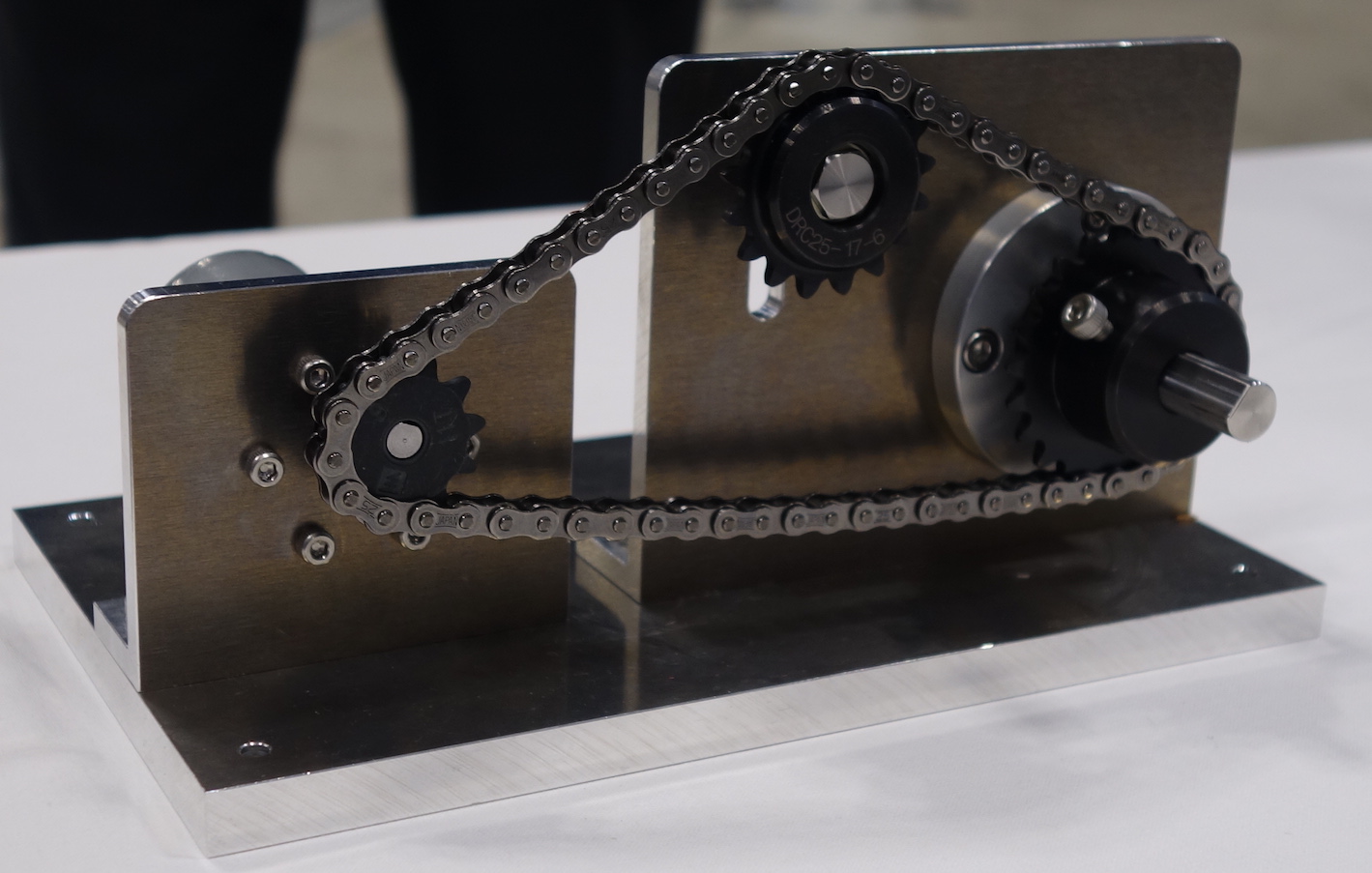}
    \caption{
        The drive units to be assembled by autonomous robot systems.
        \textbf{Left}: original belt drive unit as used in task~3 (courtesy of the WRS Industrial Robotics Competition Committee~\protect\cite{wrs2018_rulebook}).
        \textbf{Middle and right}: the surprise assemblies of task~4 with second highest and highest difficulty, respectively.
    }
    \label{fig:belt-drive-unit}
\end{figure}

The Assembly Challenge has been motivated by tasks that are typical in an industrial setting, with rules that are influenced by the constraints of industrial practice rather than by specific research challenges. 
This is in contrast to toy problems such as robot soccer~\cite{niemueller2015evaluation} or the popular FIRST competitions~\cite{buckhaults2009increasing}, which advance system integration and more general skills and interest in robotics. 

Finding the ``right'' challenges that lead to critical, fundamental advances while being industrially relevant is a hard problem in itself. 
Watanabe et al.~\cite{watanabe2017survey} provide a survey of robotic manipulation studies that have practical applications. 
We believe that although open-ended research and associated competitions will advance the state of the art, it is ultimately a practical application and the associated economical value that will lead to substantial investments in research by government and industry.
However, with manipulation requiring a combination of mechanism design, perception, and control, it is difficult to compare results, and even the most careful challenge problems might lead to unintended consequences. 
For example, the Amazon Picking Challenge~\cite{correll2018analysis} turned out to be mostly an object recognition challenge, with the grasping and manipulation challenge being negligible as the vast majority of items could be grasped using suction.
Similarly, the first grasping and manipulation challenge at IROS~\cite{sun2018robotic} has been won by a team that attached self-adhesive foam cubes to all objects, which could then be manipulated by a Rethink Robotics Baxter, a robot otherwise not precise enough to carry out these specific tasks. 
Both are examples of clever solutions that work around the manipulation problem, emphasize the need for vision, but do not lead to the desired jump innovation in robotic manipulation. 

Although robot competitions have incited teams to push their systems to remarkable heights, the knowledge they accumulated often dissipates after the end of the contest, when teams decompress from the heat of the competition, members leave the group and the remainder struggles to find a venue to publish their results. 
To our knowledge, only one team published an article presenting their system and experiences from the WRC (BerlinAUTs~\cite{Kroger.2019}). 

In this article, we review the solutions of participating teams and, through the use of a survey and first-hand experiences from six teams who co-author this work, present the lessons teams have drawn from their experience during the development as well as the competition itself.
The goal of this article is to document and open the experience that teams have gained to the wider robotic community, so as to facilitate and accelerate the development and deployment of new robotic solutions for assembly as well as designing competitions in general.

This paper is structured as follows.
First, we give an overview about the competition tasks and the results that were achieved generally and by individual teams.
Next, we present the approaches taken by different teams and their voices in more detail, based on data we collected through a survey and video data recorded on site.
Finally, we discuss trends and conclusions we drew, in particular regarding areas of research and application in the field of automated assembly that remain insufficiently resolved, and suggest possible avenues to remedy the situation.


\section{Competition Tasks and Results}

All teams had to perform four individual tasks during the competition that model a typical assembly workflow starting from arranging individual items into a kit (Section~\ref{sec:kitting}) and then assembling them. In order to sample a wide variety of skills, there were three assembly tasks: a simple 2D assembly task ("taskboard", Section~\ref{sec:taskboard}), a 3D assembly task (Section~\ref{sec:3dassembly}), and a 3D assembly task with surprise parts (Section~\ref{sec:surprise}). 
The scores from each task were combined at the end to rank order the participating teams. 
As the rule set is relatively complex, we only aim to give an overview. 
The complete rule book is available online~\cite{wrs2018_rulebook}.

\subsection{Individual Tasks}

In all tasks, a similar set of objects is used, consisting of bolts, screws/nuts (M3 to M12), pulleys and other parts required to assemble the belt drive unit depicted in Figure~\ref{fig:belt-drive-unit} on the left.
The parts can be purchased online (Misumi), so the competition tasks can be re-staged and used as a benchmark even by research groups who did not participate in the challenge.

\subsubsection{Kitting}\label{sec:kitting}
\begin{figure}[!htb]
\centering

\includegraphics[width=2in,angle=90]{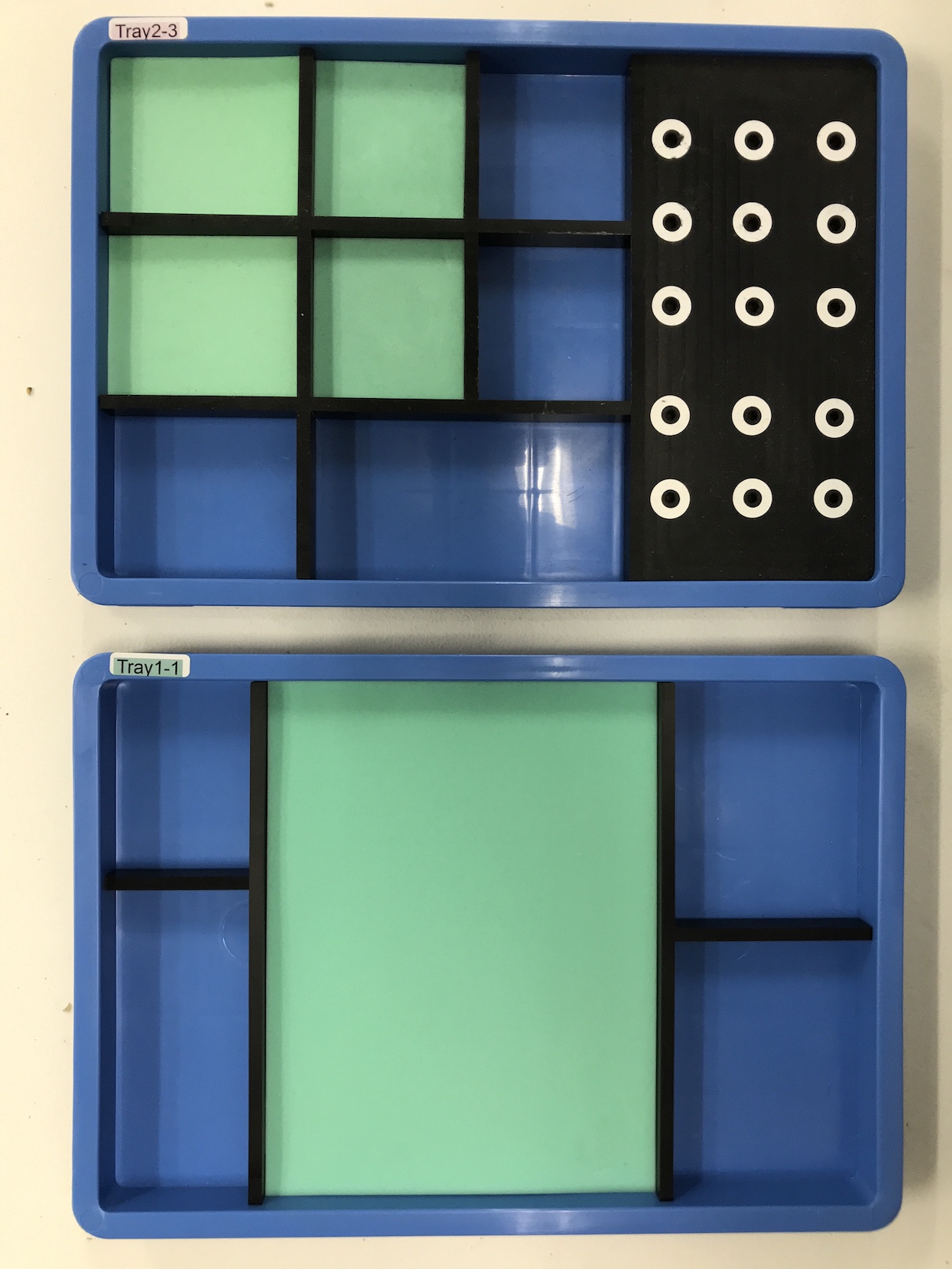}
\includegraphics[height=2in]{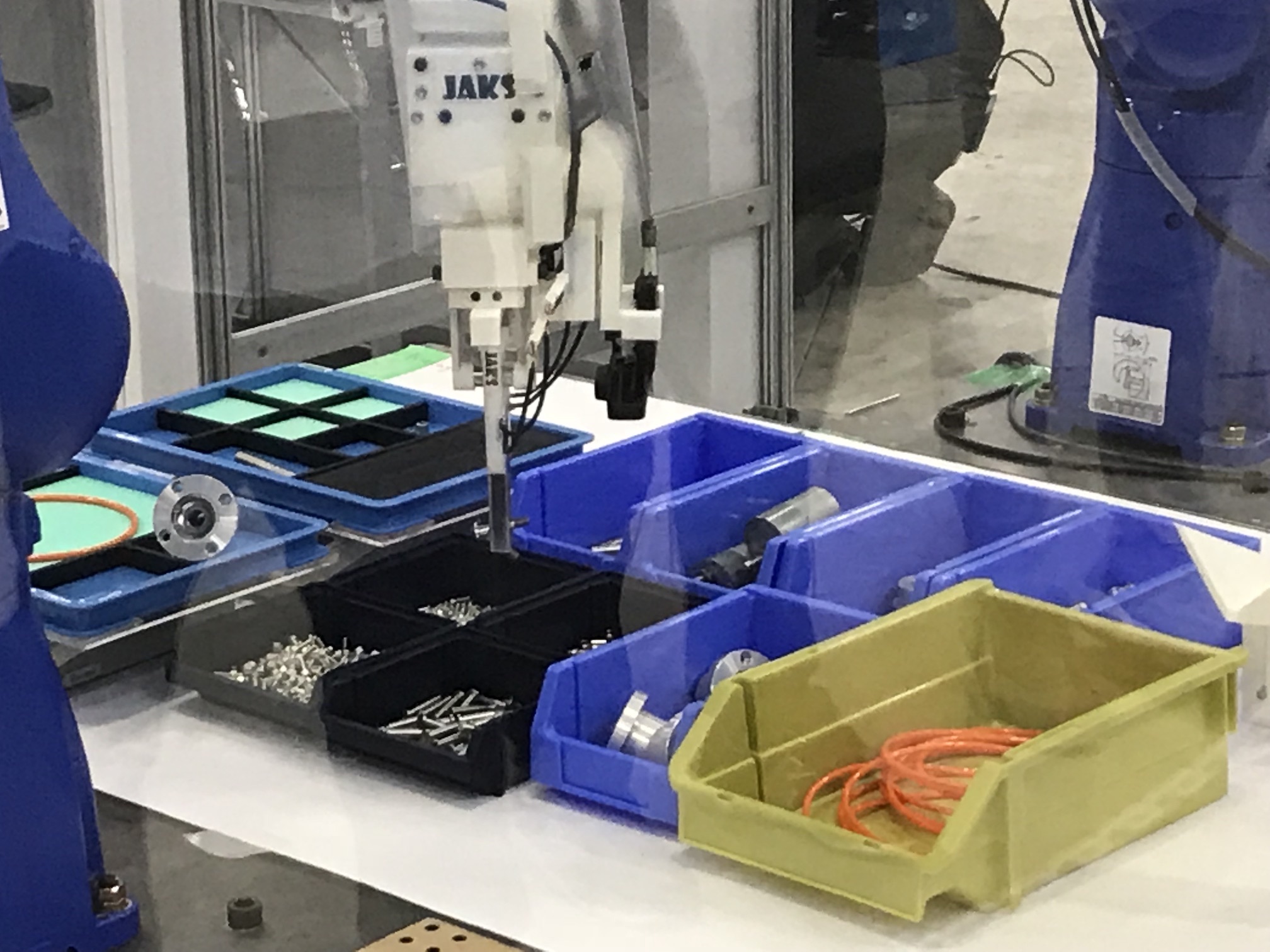}
\caption{Kitting trays used during the WRS challenge (left). Bins with bulk parts (Team JAKS).}
\label{fig:kitting_details}
\end{figure}

Parts for a manual assembly line are usually provided in a ``kit'' that provide all parts that a worker needs to complete on step in an assembly process. 
Such kits can come in various forms, providing more or less structure to a task, for example in form of a bag, a tray, a tray with compartments, or a tray where each part is arranged in a holder.
Such arrangements trade preparation time with efficiency during the assembly process and facilitate automation. 
The sample kits used during the challenge are shown in Figures~\ref{fig:kitting_details} and \ref{fig:taskboard}. 
The trays shown in Figure~\ref{fig:kitting_details} needed to be populated by a robotic system. 

The process of arranging parts in a kit is known as ``kitting'' and usually involves selecting individual items from bulk containers. For the WRS challenge, all parts were provided in bins of different sizes, such as shown in Figure~\ref{fig:kitting_details}. 
This task mimicked a ``bin picking'' scenario, which is a canonical task in industry and logistics.

However, in addition to simply picking the items, they had to be placed in the bins in the correct orientation, which required an extra degree of manipulation and planning.
For example, screws needed to be inserted heads-up into the 15 holes indicated by the white circles in Figure \ref{fig:kitting_details}, left.

\begin{figure}[!htb]
    \centering
    \includegraphics[height=1.5in]{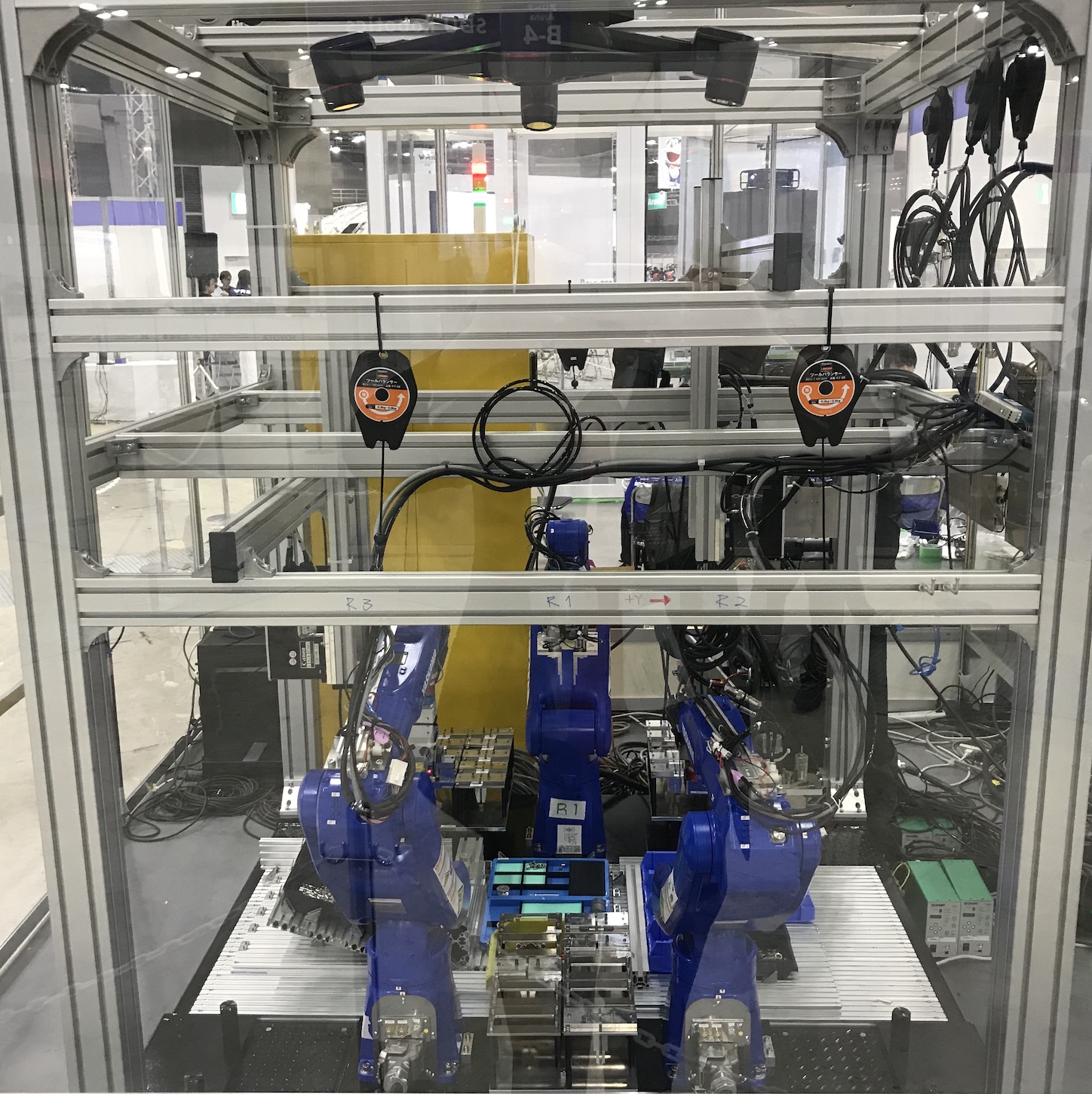}
    \includegraphics[height=1.5in]{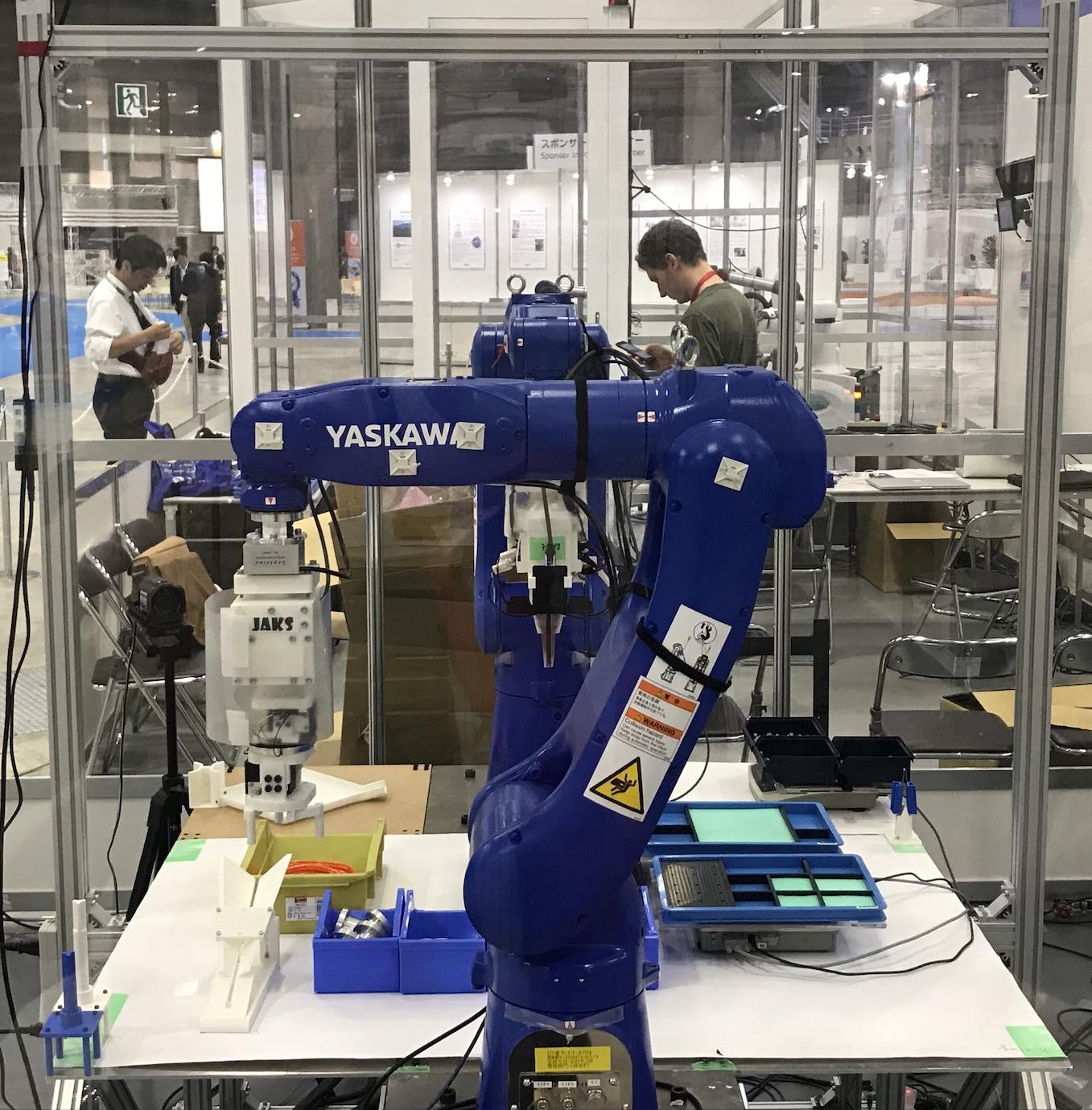}
    \includegraphics[height=1.5in]{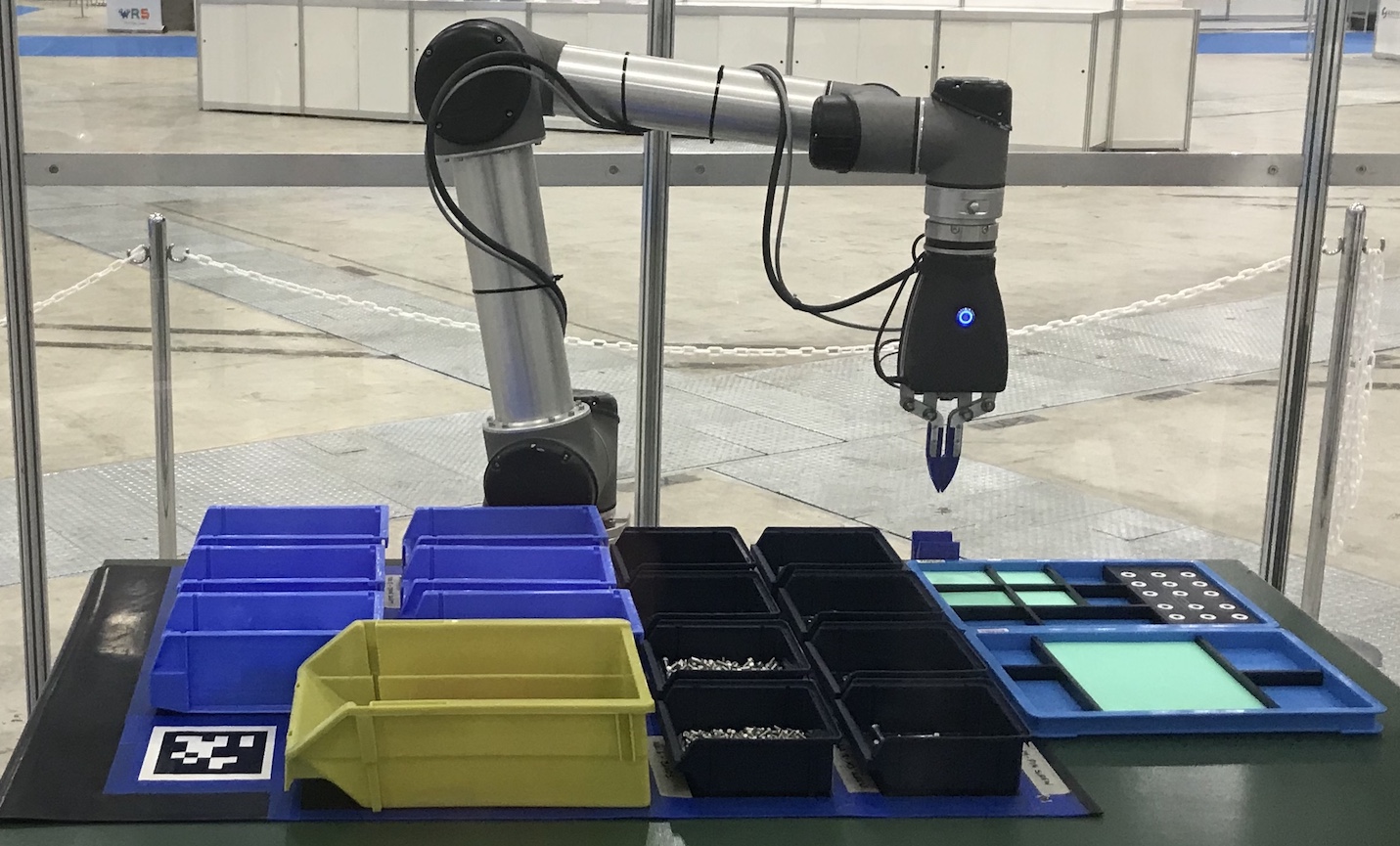}
    \caption{Sample solutions to the kitting task from the teams Robot System Integrators, JAKS, and Robotic Materials (from left to right). The different solutions illustrate the drastically varying use of external sensors and utilities such as tool changers.}
    \label{fig:kitting-systems}
\end{figure}

For each trial, 10~bins with parts were randomly chosen out of a total of 15 available parts. 
The teams were given a set list which defined the parts that had to be picked from the bins and placed in their corresponding compartment in the kitting trays.  
The set list contained three sets with ten parts each, which were to be completed within 20 minutes.
Exemplary setups for this task are shown in Figure~\ref{fig:kitting-systems}.

\subsubsection{Task Board}\label{sec:taskboard}

\begin{figure}[!htb]
    \centering
        \includegraphics[height=1.7in]{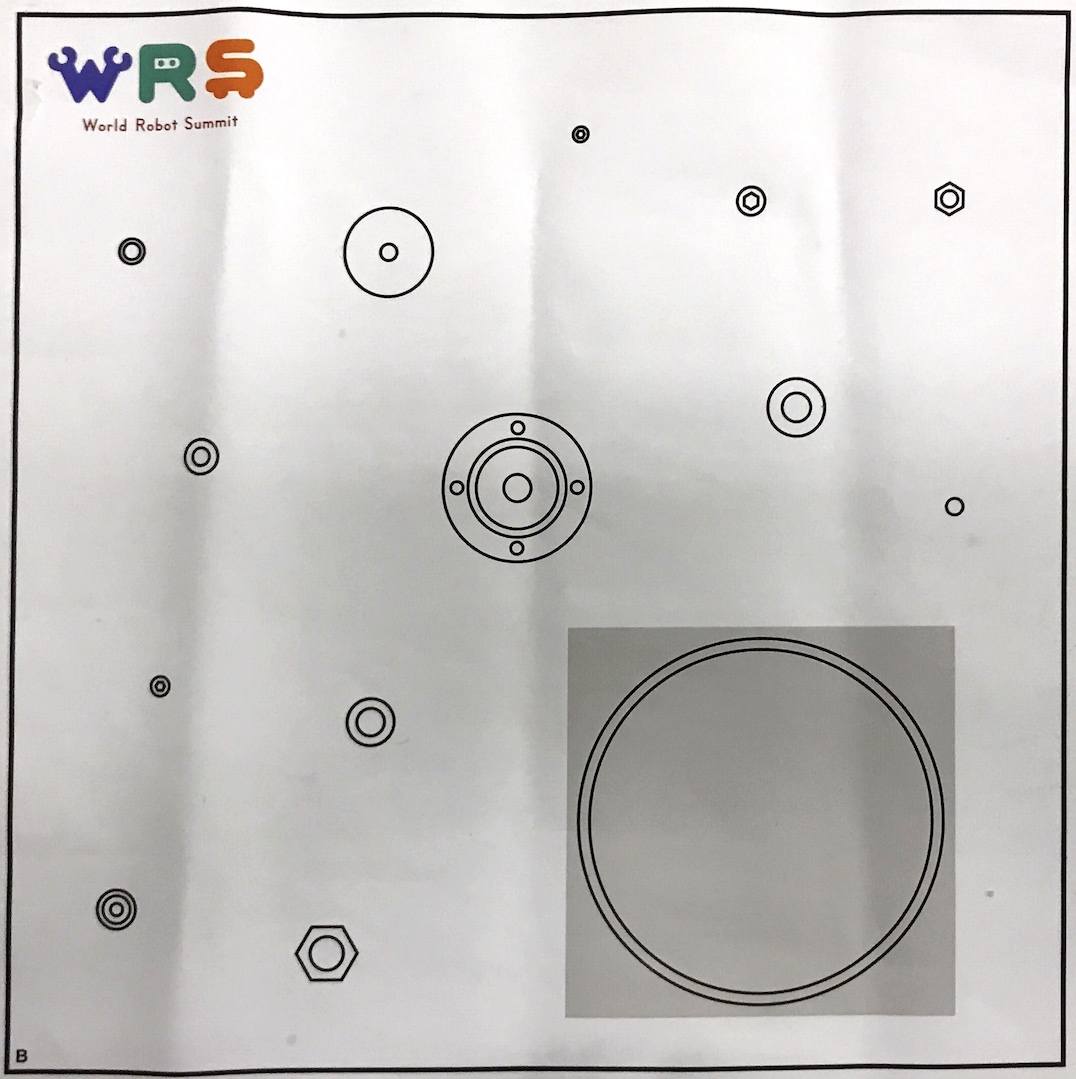}
        \includegraphics[height=1.7in]{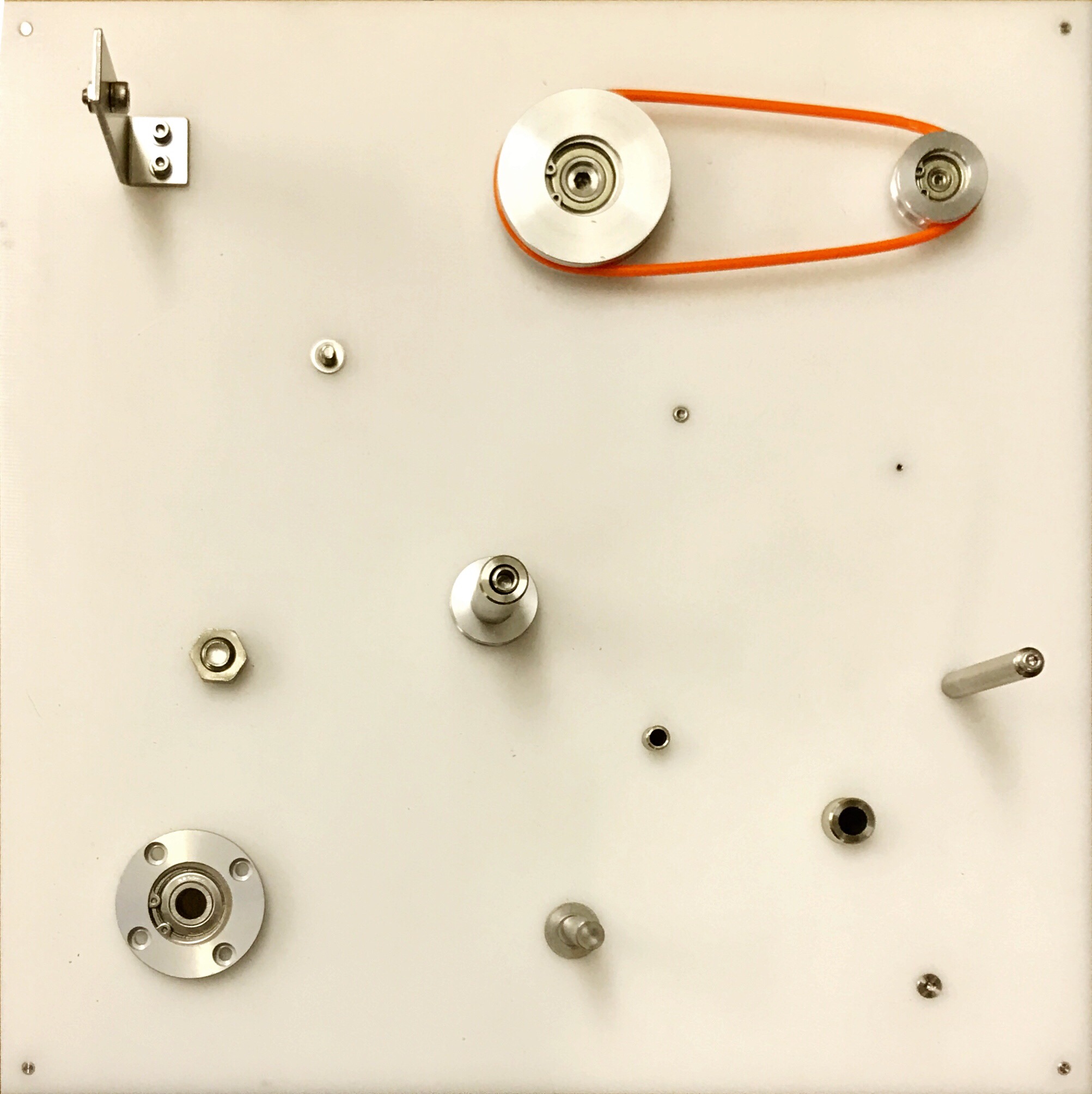}
        \includegraphics[height=1.7in]{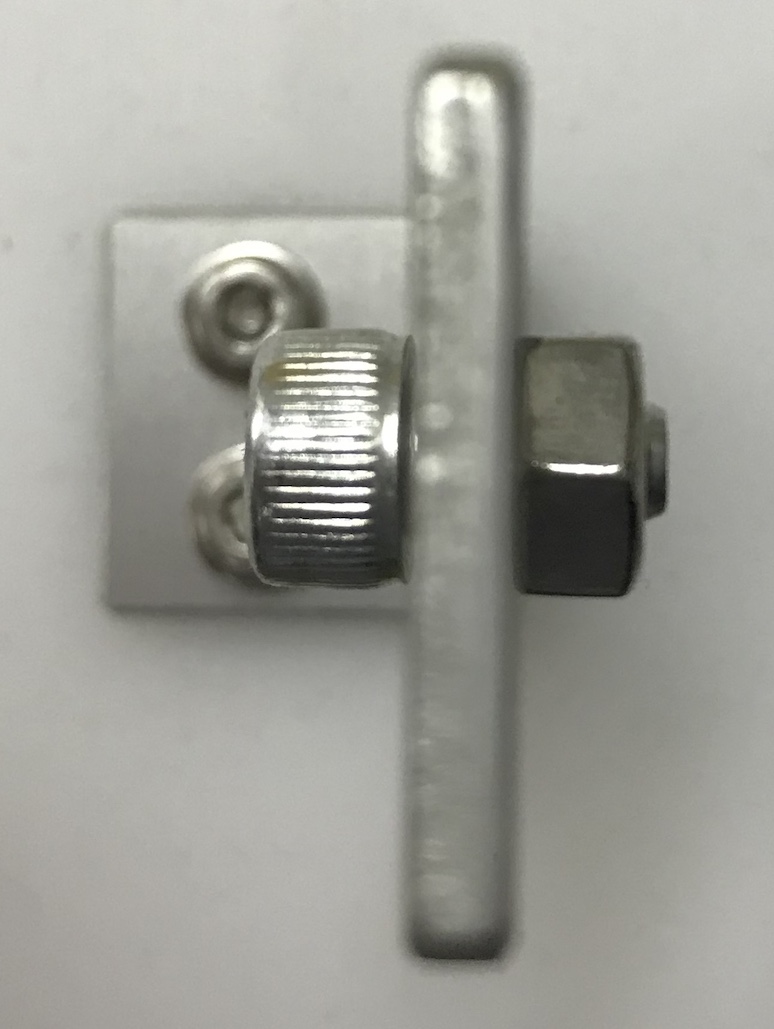}
    \caption{Kitting mat provided by the WRS organizers (left), completed task board (center), and close-up (top-view) on the horizontal insertion task.}
    \label{fig:taskboard}
\end{figure}

The first assembly task aims at testing the basic technologies required for robotic assembly.
14~parts had to be picked from a placement mat (Figure \ref{fig:taskboard}, left) and assembled onto a task board (Figure \ref{fig:taskboard}, center). 
All parts and sub-assemblies were industrial grade, meaning that fits were tight and required non-trivial contact forces, including slight stretching of the rubber band.

The placement of the parts on the mat was not arbitrary: The allowed contact surfaces and parts' orientations were pre-defined, and the layout of the mat could differ, so the position of each part had to be detected or taught during each trial. 

Most assembly operations were in the manner of ``insertion into a hole'', ``screwing into a tapped hole'', or ``fasten nut and bolt''. 
The majority of the tasks required vertical insertions, except for inserting a screw into the angle shown in the top left of Figure~\ref{fig:taskboard}, center. 
In addition to being inserted, a nut needed to be mounted from the other side, so that both parts had to be held simultaneously. 
This is illustrated in the close-up (top-view) in Figure \ref{fig:taskboard}, right. 
Each operation was scored according to its completion level, to acknowledge when e.g. a bolt and nut have contact, but are not completely fastened. 
As the task board was meant to test the robot system's capability to perform each subtask, it came with some parts pre-assembled to the board, such as two pulleys for the belt assembly, and an angle for the horizontal nut and screw fastening.
Please refer to the complete WRS manual for details on the task board setup. 

\subsubsection{Assembly}\label{sec:3dassembly}

\begin{figure}[!htb]
    \centering
    \includegraphics[height=1.55in]{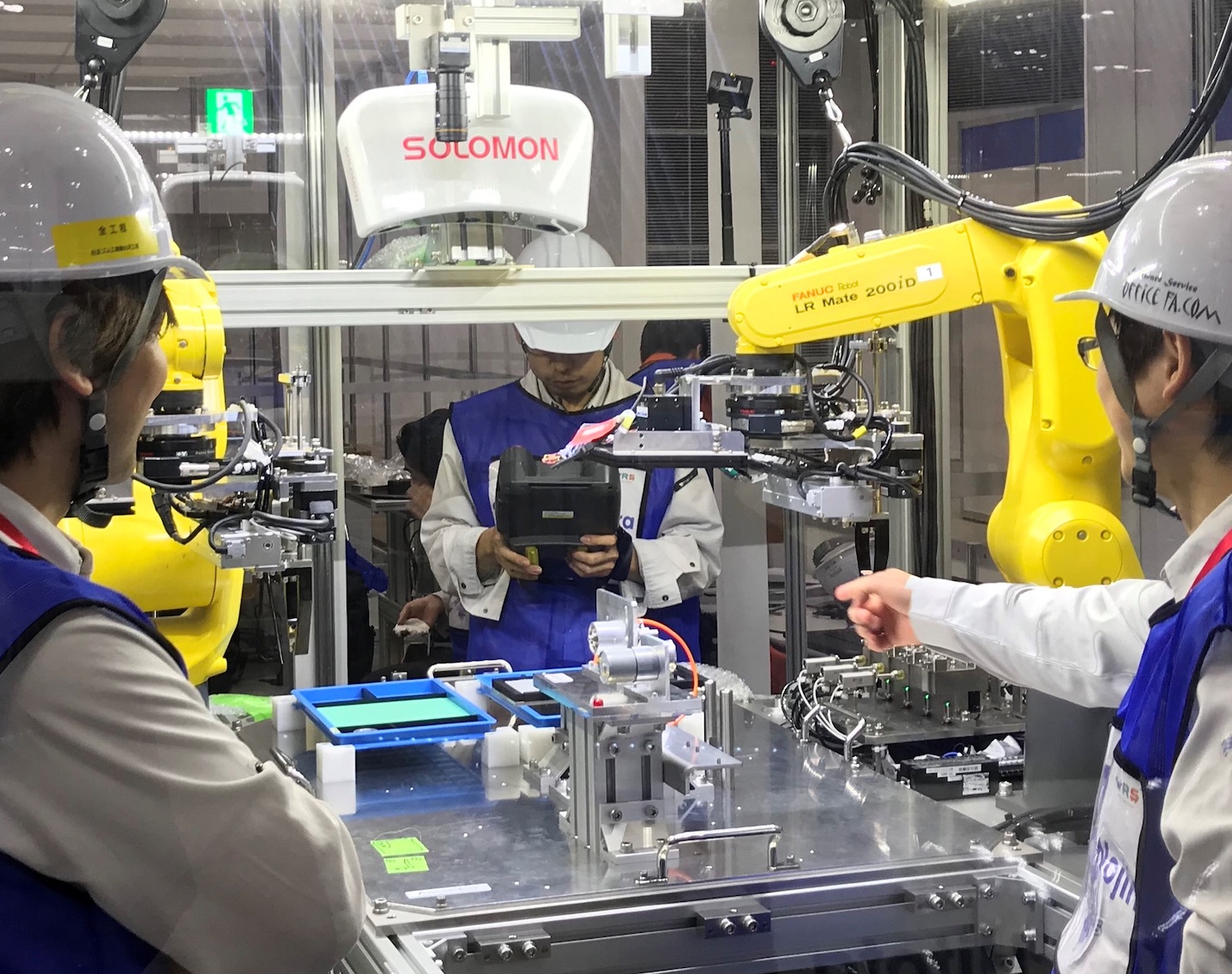}
    \includegraphics[height=1.55in]{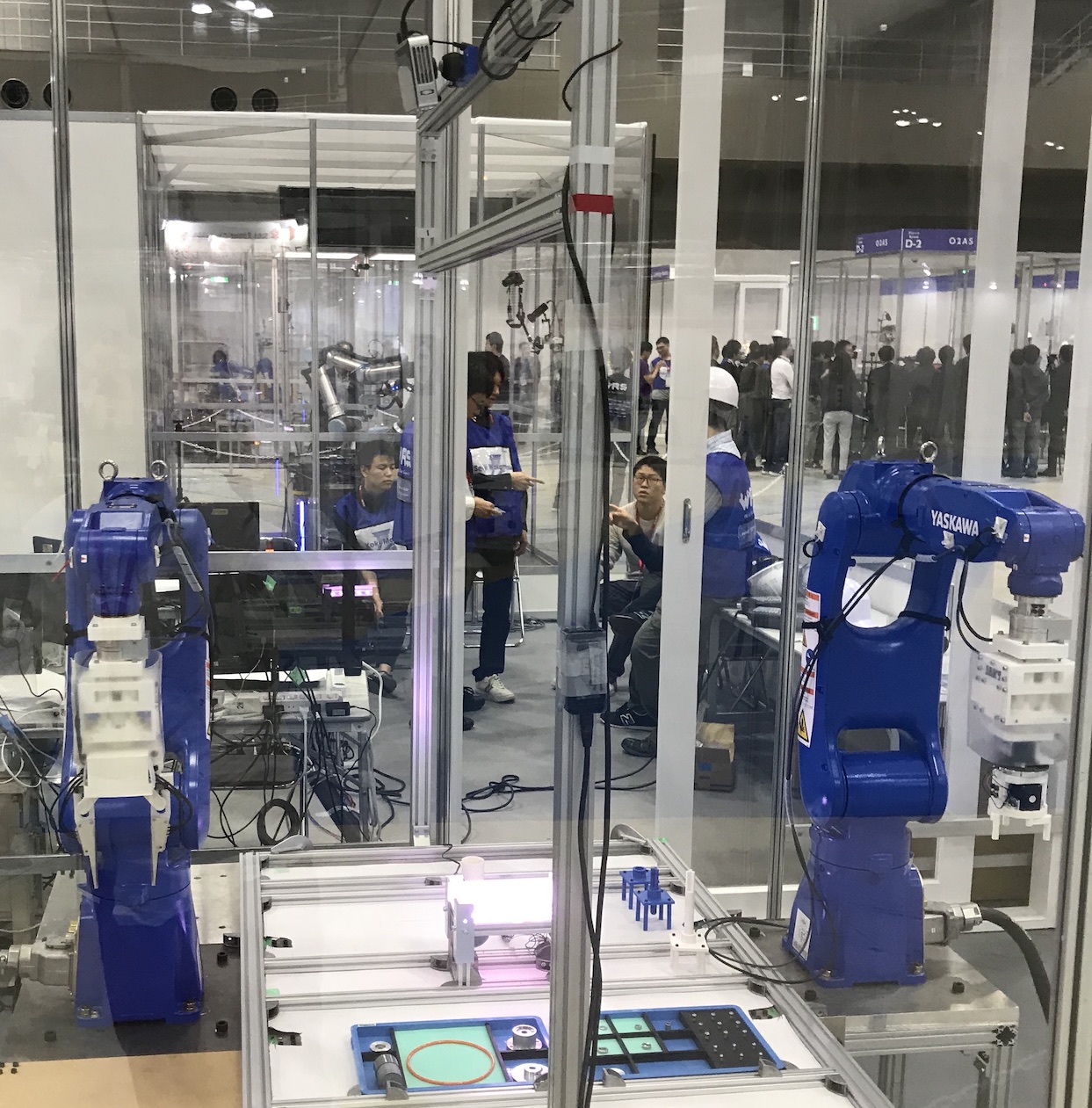}
    \includegraphics[height=1.55in]{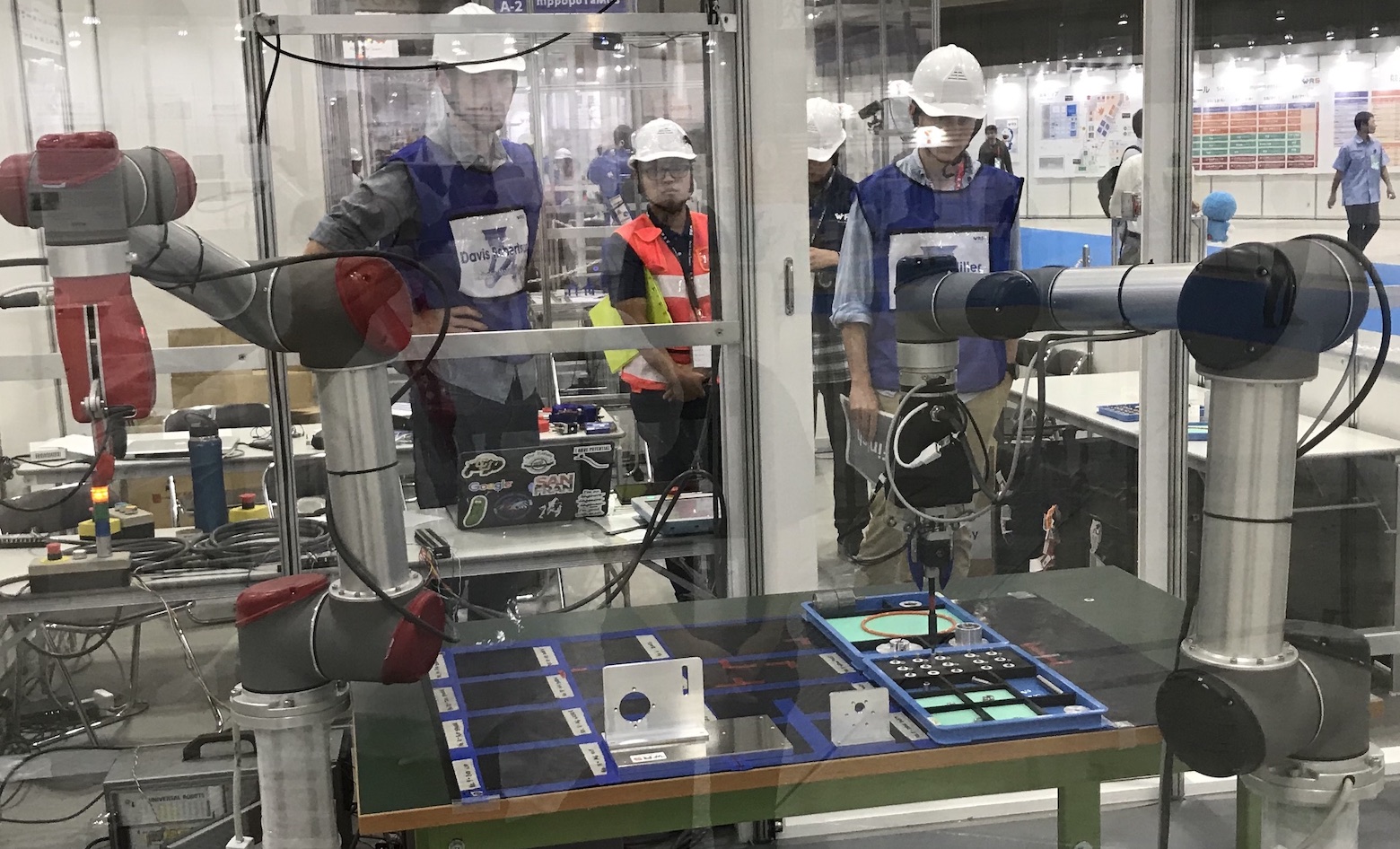} \\
    \includegraphics[height=1.5in]{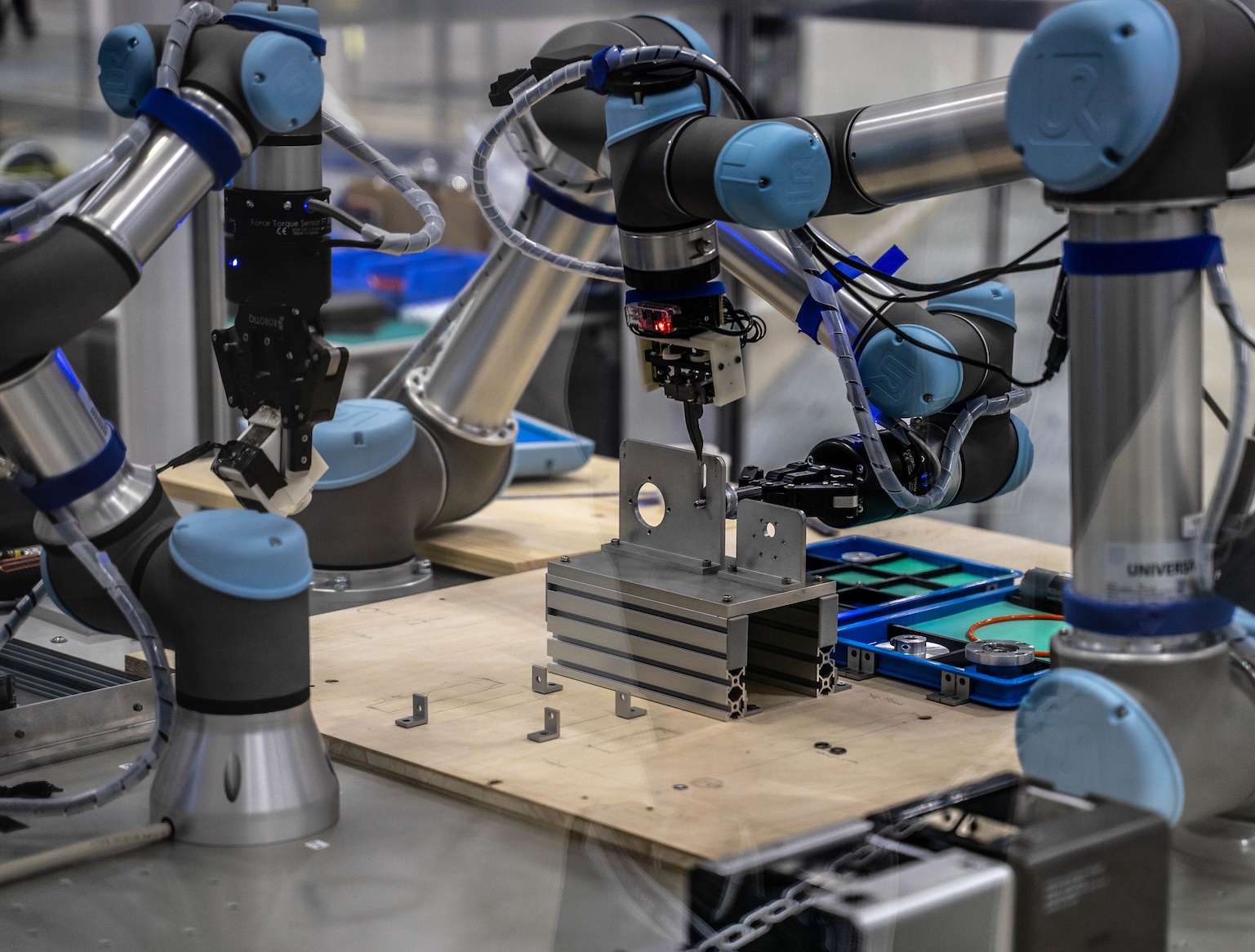}
    \includegraphics[height=1.5in]{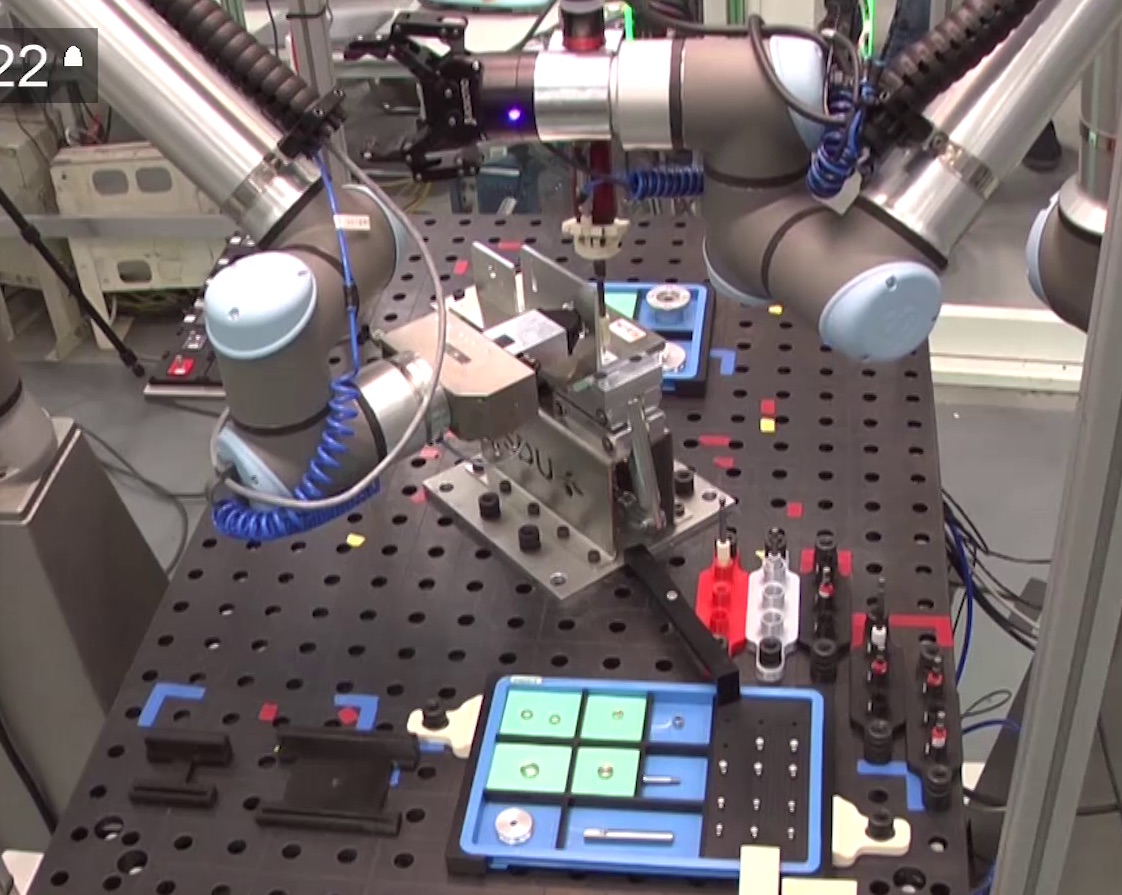}
    \includegraphics[height=1.5in]{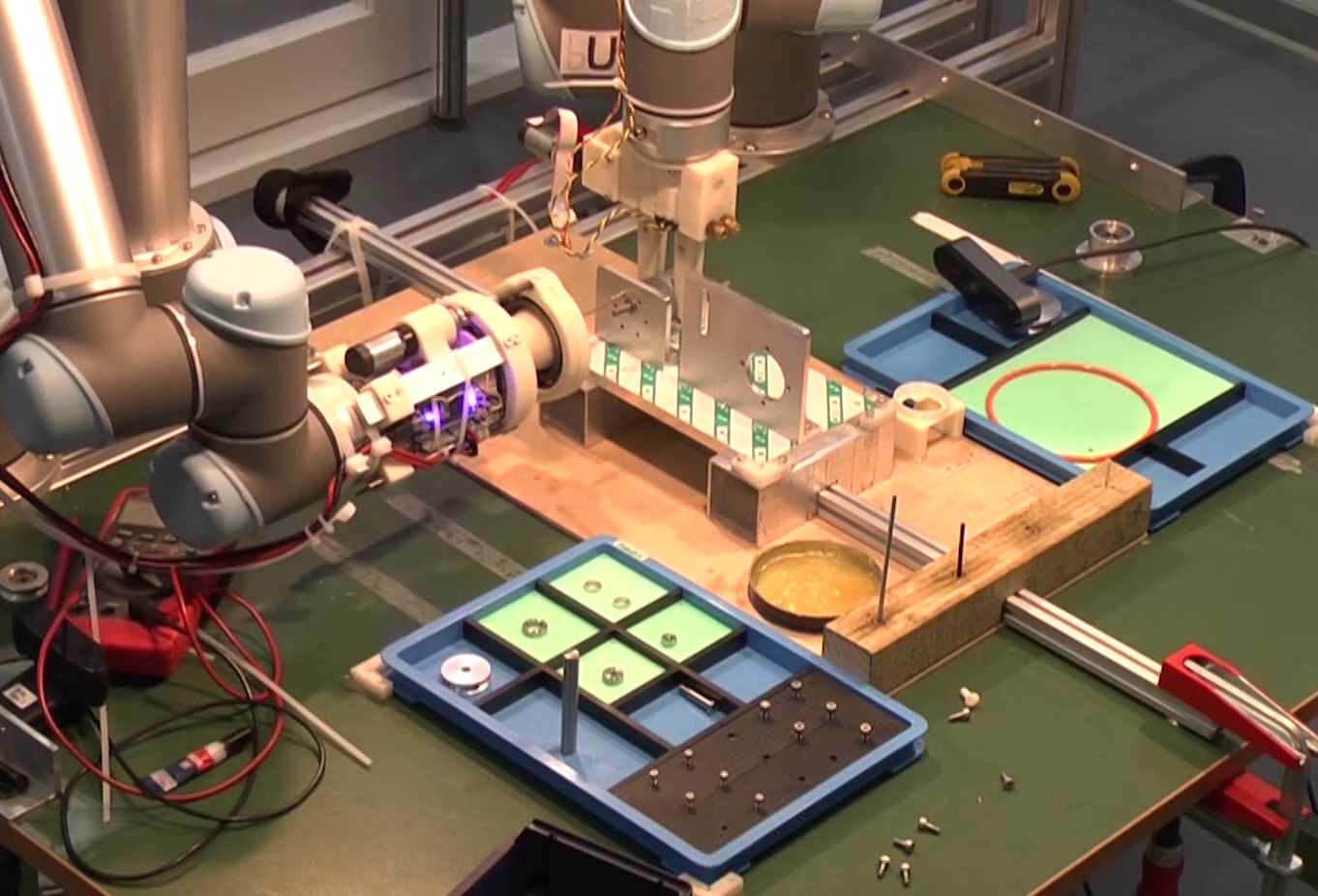}
    \caption{Sample solutions to the 3D assembly task from the teams FA.COM, JAKS, Robotic Materials (left to right, top), O2AS, SDU Robotics, Cambridge Robotics (bottom). The different solutions illustrate the drastically varying use of external sensors and utilities such as tool changers.}
    \label{fig:3dassembly}
\end{figure}

The most prominent task of the competition was the assembly of the belt drive unit, depicted in Figure~\ref{fig:belt-drive-unit}, which is representative of sub-assemblies within a complex mechatronic product such as a laser printer, dishwasher, or AC system. 
All required parts (except for the base plates) were arranged in the kitting trays (Figure \ref{fig:kitting_details}) that were also used for the previous task. 
The base plates instead could be freely positioned by each team, including . 
The complete assembly was divided into eight subtasks, which were scored according to their level of completion.
Teams could reset individual subtasks during the competition, as long as they were started (e.g. at least one part used in the subtask has been moved by the robot system).
See Figure~\ref{fig:3dassembly} for sample setups of the assembly task.

\begin{figure}[!htb]
    \centering
    \includegraphics[height=1.7in]{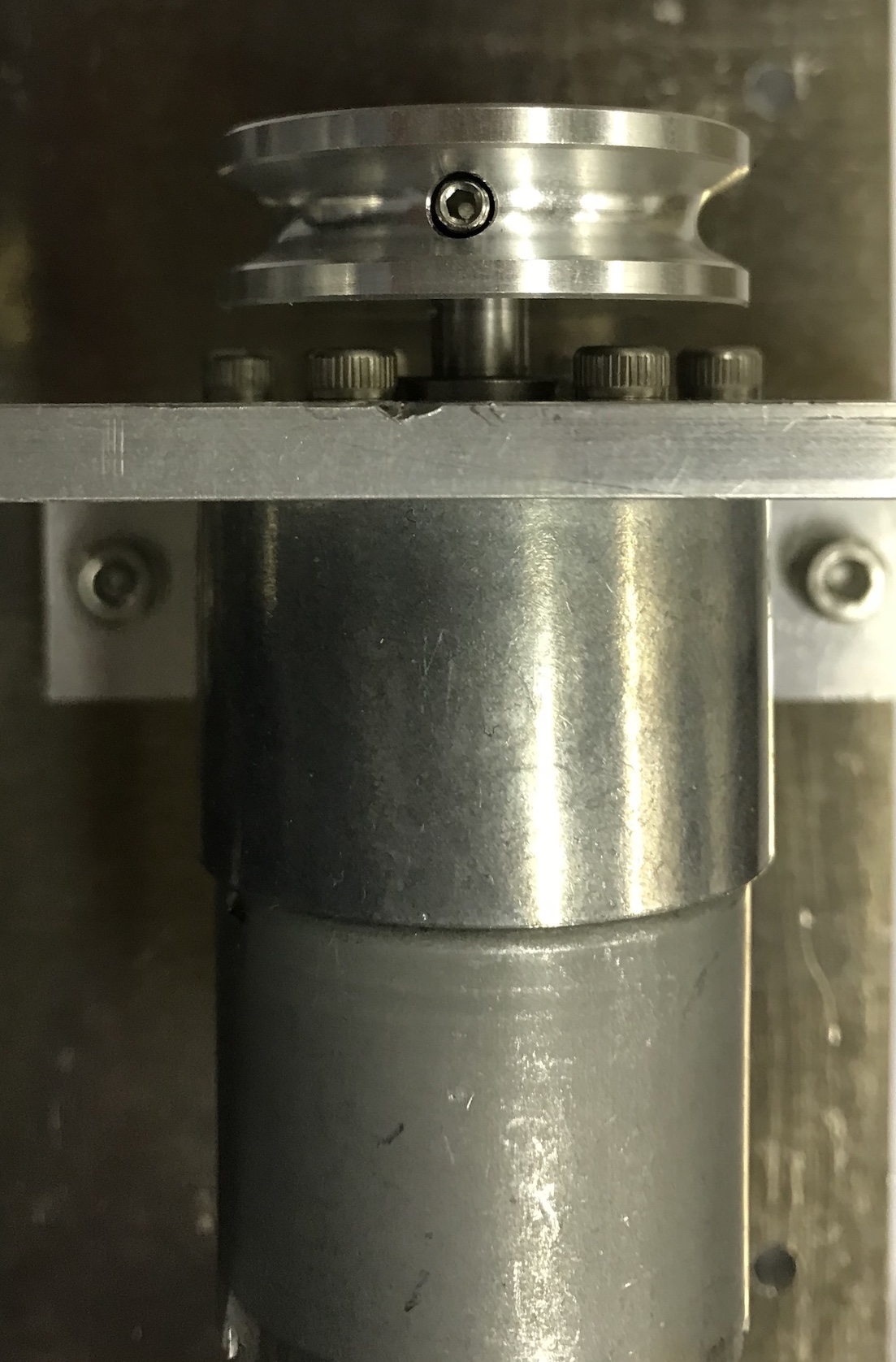}
    \includegraphics[height=1.7in]{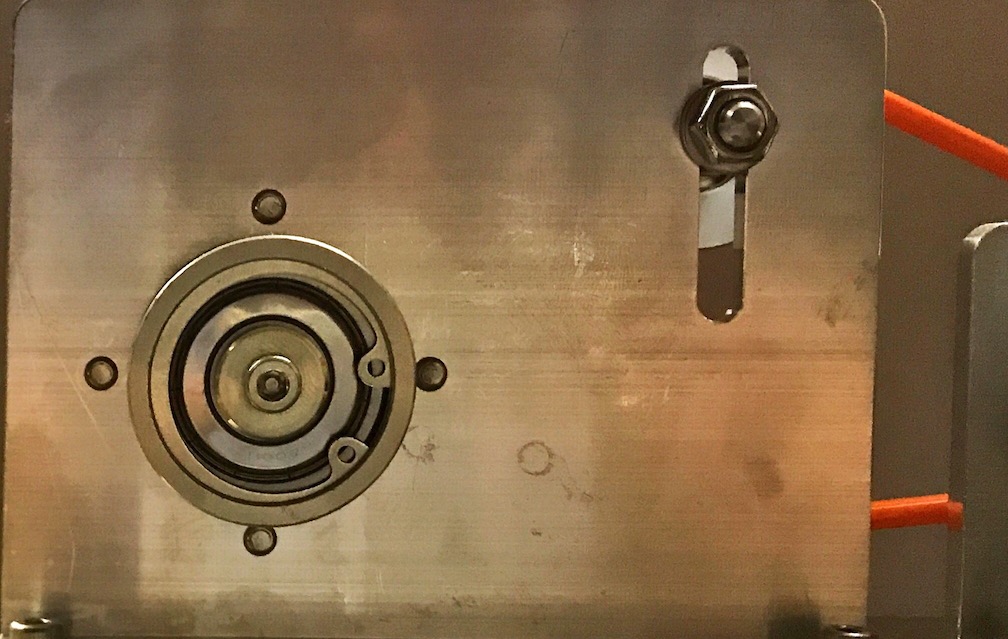}
    \includegraphics[height=1.7in]{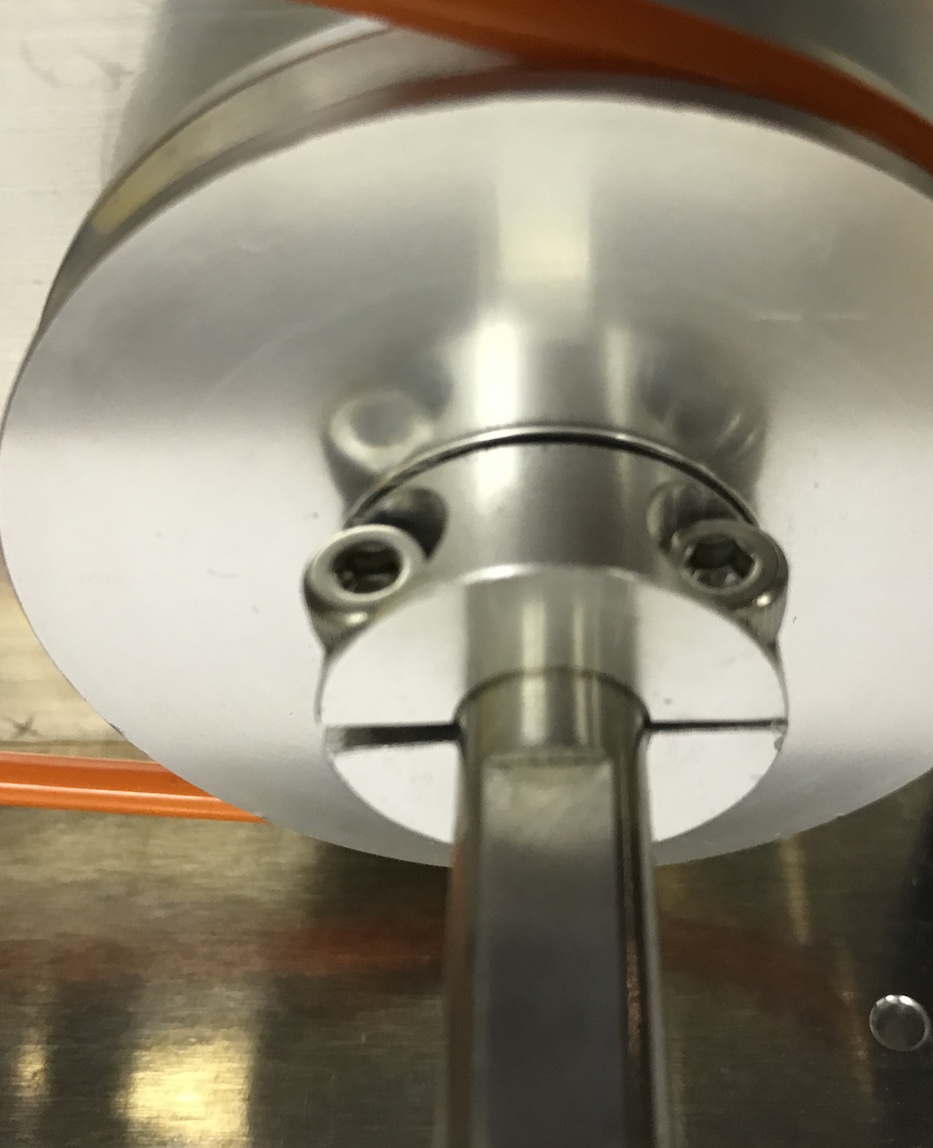}
    \caption{
    Close-up views of specific challenges in the 3D assembly problem. 
    \textbf{Left}: (seen from above) the motor axis needed to be inserted into the mounting plate so that the six screw holes are aligned with the holes in the mounting plate. Note that the motor axis was not in the center of the motor case.
    \textbf{Center}: (seen from behind) inserting the ball bearing into the mounting plate required considerable force (tight fit), as well as alignment of four screw holes that are inserted from the other side. This close-up also shows the sliding mount for the pulley that stretches the rubber band.
    \textbf{Right}: To attach the pulley to the shaft, it had to be held in place while the two screws were fastened. 
    }
    \label{fig:3dassembly_challenges}
\end{figure}

Due to the low tolerances of the parts, their insertion and assembly required non-trivial contact forces. 
While the precise tolerances of the parts were not given, they were such that some insertion operations were not trivial even for human operators (due to jamming).
Some parts also required the alignment of screw holes, such as the motor and ball bearing assemblies, as well as the manipulation of multiple parts at once. 
These cases are illustrated in Figure \ref{fig:3dassembly_challenges}. 

During the competition, 20 minutes were allocated for the assembly of one belt drive unit.
As a reference value: Unskilled human test subjects take around 6 to 8~minutes to assemble one belt drive unit, and around 4~minutes after some practice.

\subsubsection{Surprise Assembly}\label{sec:surprise}
The fourth and last task is an extension of the assembly task.
According to the scores achieved in the previous tasks, teams could choose among four difficulty levels.
The higher the difficulty level, the more surprise parts were part of the assembly, which were unknown until the day before the trial.
For example, in the highest difficulty level the belt has been replaced by a chain (see Figure~\ref{fig:belt-drive-unit} middle and right) and the pulleys by gear sprockets, which required completely different manipulation strategies.
This task particularly targets ``level 5'' automation, as the specifications of the surprise parts were available only the day before, and the physical parts were provided only two hours prior to the trial.
Note that during their trials, teams had to not only assemble the surprise drive unit but also the original belt drive unit, to prove their flexibility to switch between different variants.

\subsection{Competition Procedure}

In advance to the actual competition, the teams had two days for setting up their systems at the venue.

All teams had to pass a safety check before being allowed to operate their systems. 
In particular, all systems needed to be connected to an emergency stop button, a feature that is standard in industrial robotic control systems, but requires additional engineering for custom solutions. 

Each subsequent day was dedicated to one particular task.
While each team had two trials for the taskboard, kitting and assembly task, there was only one trial for the surprise assembly.
The best score of both trials counted, and the total sum determined the ranking.
Furthermore, a small amount of points (less than 6\% of the top 5 teams' scores) was given based on a technical evaluation of the system by the judges.

\subsection{Competition Results}
Altogether, 16~teams competed during the challenge, around half of the teams from Japan and the other half from the rest of the world. The WRS organizers provided generous stipends to support travel and shipping cost.

Team SDU Robotics from the University of Southern Denmark won the competition ahead of team JAKS and FA.COM from Japan on second and third place, respectively.
Teams SDU Robotics and O2AS (4th place) also received special awards from the Robotic Society of Japan (RSJ) and the Society of Instrument and Control Engineers (SICE) for their solutions. Team Cambridge Robotics was also awarded the Innovation Award, and Team Robotic Materials the President's award by the Japanese Society of Mechanical Engineers (JSME).

\begin{figure}[!htb]
    \centering
    \includegraphics[width=.9\columnwidth]{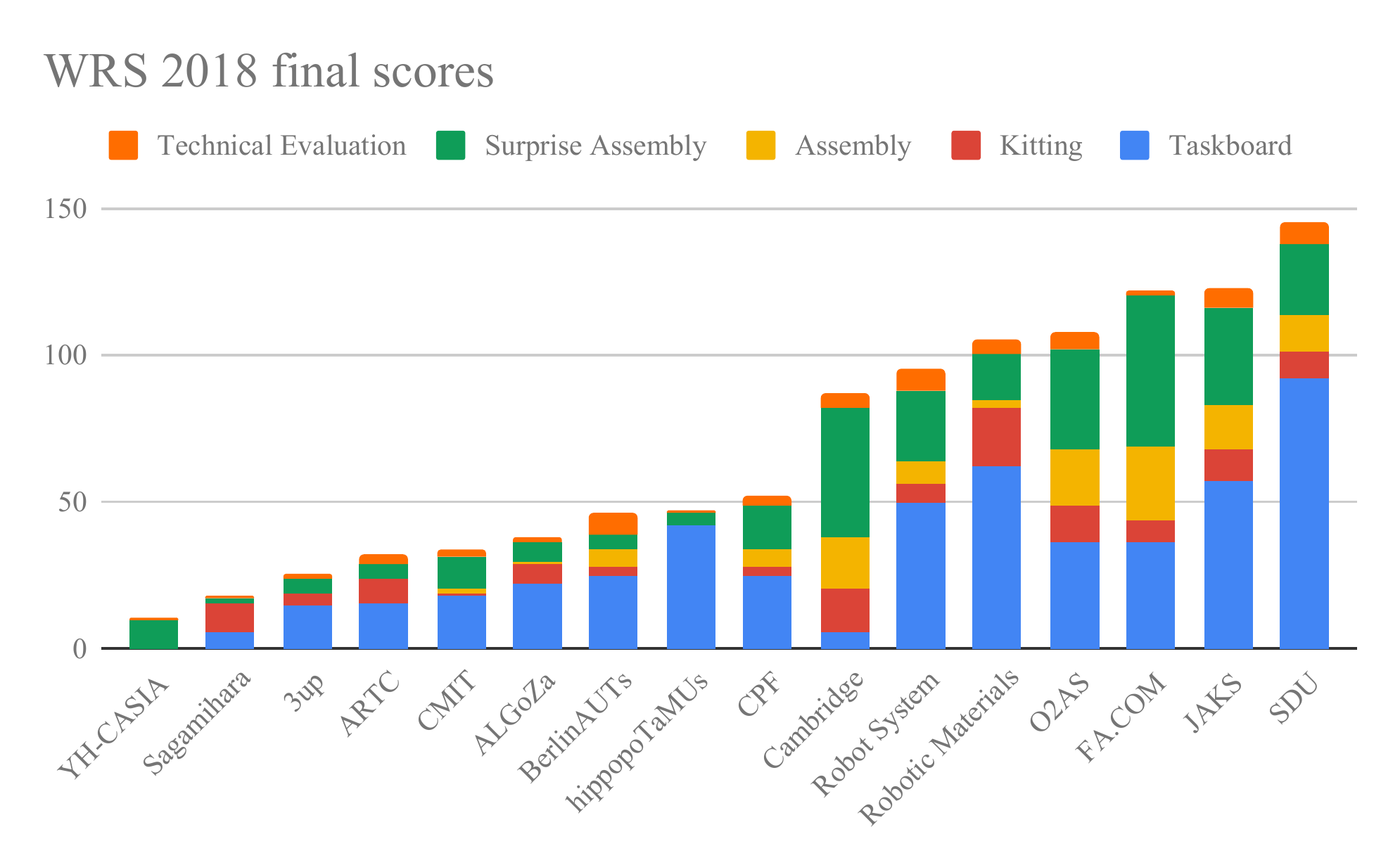}
    \caption{
        Scores of all teams, separated by task.
    }
    \label{fig:wrs-scores}
\end{figure}

The achieved scores are illustrated in Figure~\ref{fig:wrs-scores} and reveal the tasks which gave the most points.

Interestingly, none of the tasks could be completely solved by any of the teams.
While SDU Robotics gained most of their points by almost finishing the taskboard task, FA.COM were the only team throughout the competition to once successfully assemble the belt drive unit. 
We also note that the overall ranking is not representative for performance in individual sub-tasks. 
For example, the Cambridge (7th overall) approach appeared to be very adept at reconfiguration (second best in the surprise assembly task), with Robotic Materials (5th overall) performing best at the kitting task, and FA.COM (3rd overall) obtaining the highest score in the assembly category.

Comparing the results from each task, we observe that the teams scored most of the points during the taskboard task, followed by the surprise assembly.
This might be surprising, as the surprise assembly is the hardest task.
However, it also contained the assembly of the original belt drive unit which had been practiced in the two trials the day before, and the easier surprise assemblies did not require too many modifications.
The points gathered in this task might therefore not necessarily originate from assembling previously unknown parts, but rather from assembling the standard belt drive unit.
As such, FA.COM performed the successful assembly of the belt drive unit during their surprise assembly trial.
The assembly took 50 minutes and 7 restarts (due to a dropped screw and failures in the pulley/belt assembly), but less than 20 minutes the day after, during the exhibition --- a testament to the repeatability of current robotic systems once the process is set up.

Most of the teams performed relatively poorly on the kitting task.
In the symposium immediately following the competition, teams expressed that this may have been due to the limited development time and the different skill set required for the kitting task.


\section{Survey and Comparisons}

The field of participants in WRC 2018 was rather diverse, with teams from universities as well as from companies, all with different expectations and possibilities to face the competition.
Accordingly, the systems and strategies in WRC 2018 were demonstrating a wide range of solutions as well as some common denominators.   

\subsection{Overview and Methodology}

To compare the different approaches taken by the teams, we have employed the following data collection strategies. 
After the competition, we have conducted a survey among the teams. 
The survey was implemented in Google Forms and was distributed via email as well as in the Facebook group that was used for communication between committee and teams. 
The survey consisted of 24 questions provided in English and Japanese language. 
The answers were given in multiple-choice and/or free text format.

\begin{enumerate}
 \item \textbf{Team composition}
 \begin{enumerate}
    \item Did you participate in previous competitions?
    \item How many people were involved in the development of the WRC 2018 system? (Breakdown by academic degree)
    \item How many institutions were involved in the development of your system? (Breakdown by type of institution)
 \end{enumerate}
 \item \textbf{Development costs}
 \begin{enumerate}
     \item Did you develop your system from scratch?
     \item What are the estimated hardware costs of the complete system?
     \item what is the estimated development time in person months?
     \item Do you plan to continue using the system after WRC 2018?
 \end{enumerate}
 \item \textbf{System description}
 \begin{enumerate}
     \item Please specify the number and type of robots you used.
     \item What type of grippers did you use?
     \item How many tools or fingertips did you use?
     \item If you used jigs, how many of what kind?
     \item What number and type of vision sensors did you use for which problems?
     \item What type of force control sensor did you use, and for which operations?
     \item Did you use other sensors, and what for?
     \item Which software framework did you use?
     \item What type of motion planner did you use?
     \item What type of simulation tools did you use?
 \end{enumerate}
 \item \textbf{Competition conditions}
 \begin{enumerate}
    \item Please estimate you efforts for each subtask as a percentage.
    \item Did you change the system setup in between tasks, and what for?
    \item Could you briefly describe your strategy for the assembly with surprise parts?
    \item Can you describe your overall strategy?
    \item Which components of your system caused most trouble during the competition?
 \end{enumerate}
 \item \textbf{Takeaways}
 \begin{enumerate}
    \item What is your most important lesson learned from the participation in WRC 2018?
 \end{enumerate}
\end{enumerate}

\subsection{Results}

Out of the 16 teams in WRC 2018, 9 teams took part in the survey. 
The following results are compiled from their survey responses.

\subsubsection{Team composition}
Of the responding teams, 4 were from academic institutions, and 4 from companies or mixed institutions (with both public and private R\&D obligations). 
Team 3up labelled itself a "Garage" team as they are enthusiasts who worked on their WRC 2018 system in their free time.

6 out of 9 teams had experiences from previous competitions. SDU Robotics, BerlinAUTs and 3up have participated in a competition for the first time. 
Of the academic and mixed institutions, most teams were a mix of professors working with their students --- often with support from additional engineers. 
The companies relied on their employed engineers to take part in the competition. 
Across the responding teams, the typical team size was about 10 persons (including persons supporting the development in the background). 
Team O2AS listed 27 members, SDU Robotics 18 members.
Smaller teams had 3-4 members.

\subsubsection{Development costs}

All but team CPF started their system developments for WRC 2018 from scratch, and all responding teams are positive about using their systems in the future (in projects and developments).

\begin{figure}[!htb]
    \centering
    \includegraphics[width=.9\columnwidth]{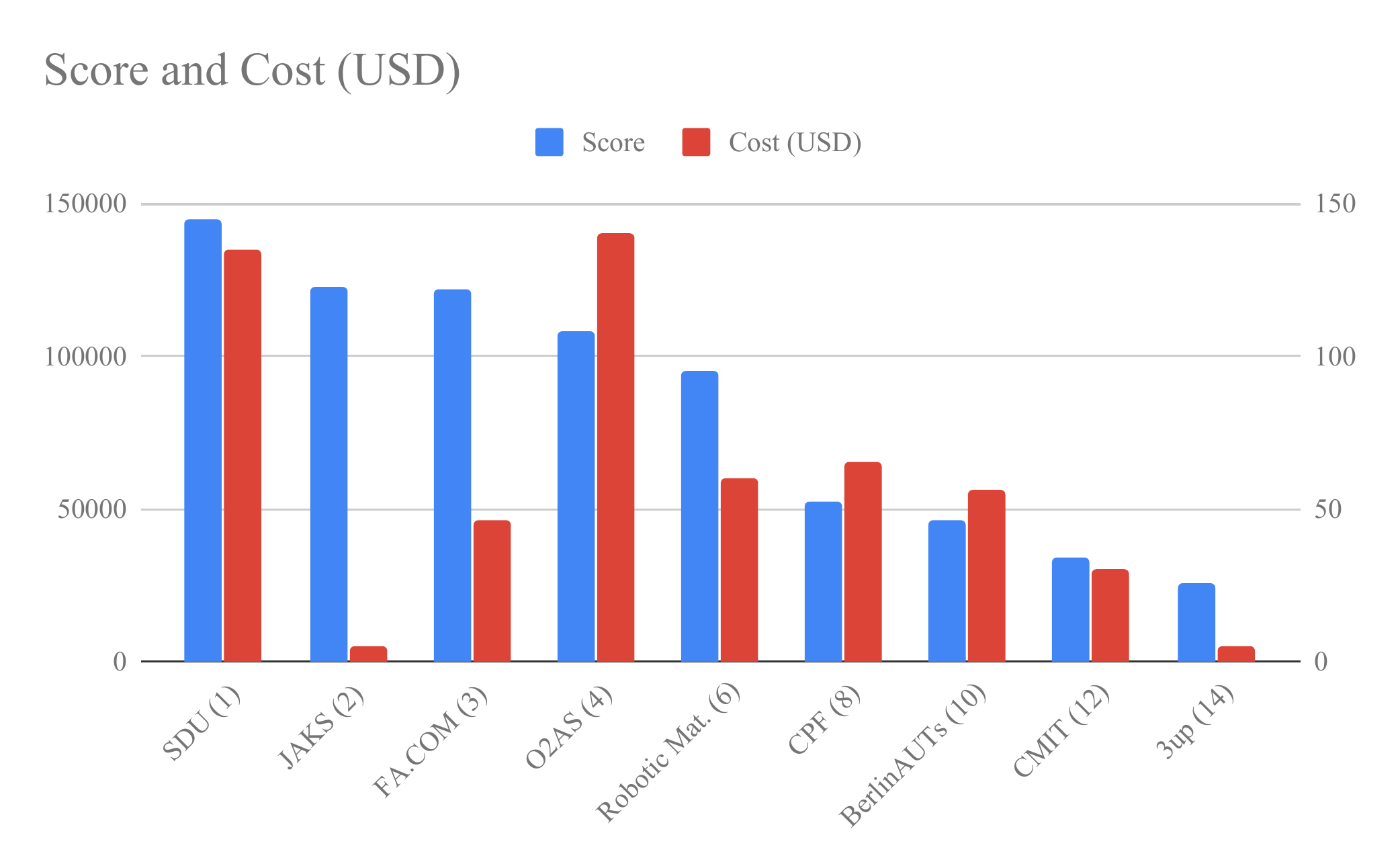}
    \caption{
        Estimated system costs, sorted by score achieved by the team. Note that some teams such as JAKS, FA.COM and 3up did not take the costs of the robots into account, as they were borrowed for the competition.
    }
    \label{fig:survey_costsvsscore}
\end{figure}

Figure~\ref{fig:survey_costsvsscore} depicts the estimated costs for the systems of the responding teams (in USD). 
Teams SDU Robotics and O2AS both listed their systems to costs about 130-140,000 USD, for instance in the case of SDU Robotics taking into account the costs for two robots, various tools and sensors, industrial worktable and PLCs. 
In direct comparison with their scores, most teams yield a ratio of about 0.8 - 1.2 points per 1,000~USD system costs. 
Exceptions are the cases of O2AS (0.77) and Robotic Materials (1.58). 
Other outliers such as JAKS, FA.COM and 3up result from teams not taking the costs for the robots into account, as some Japanese robot manufacturers offered their robots to be borrowed for the competition.

\begin{figure}[!htb]
    \centering
    \includegraphics[width=.6\columnwidth]{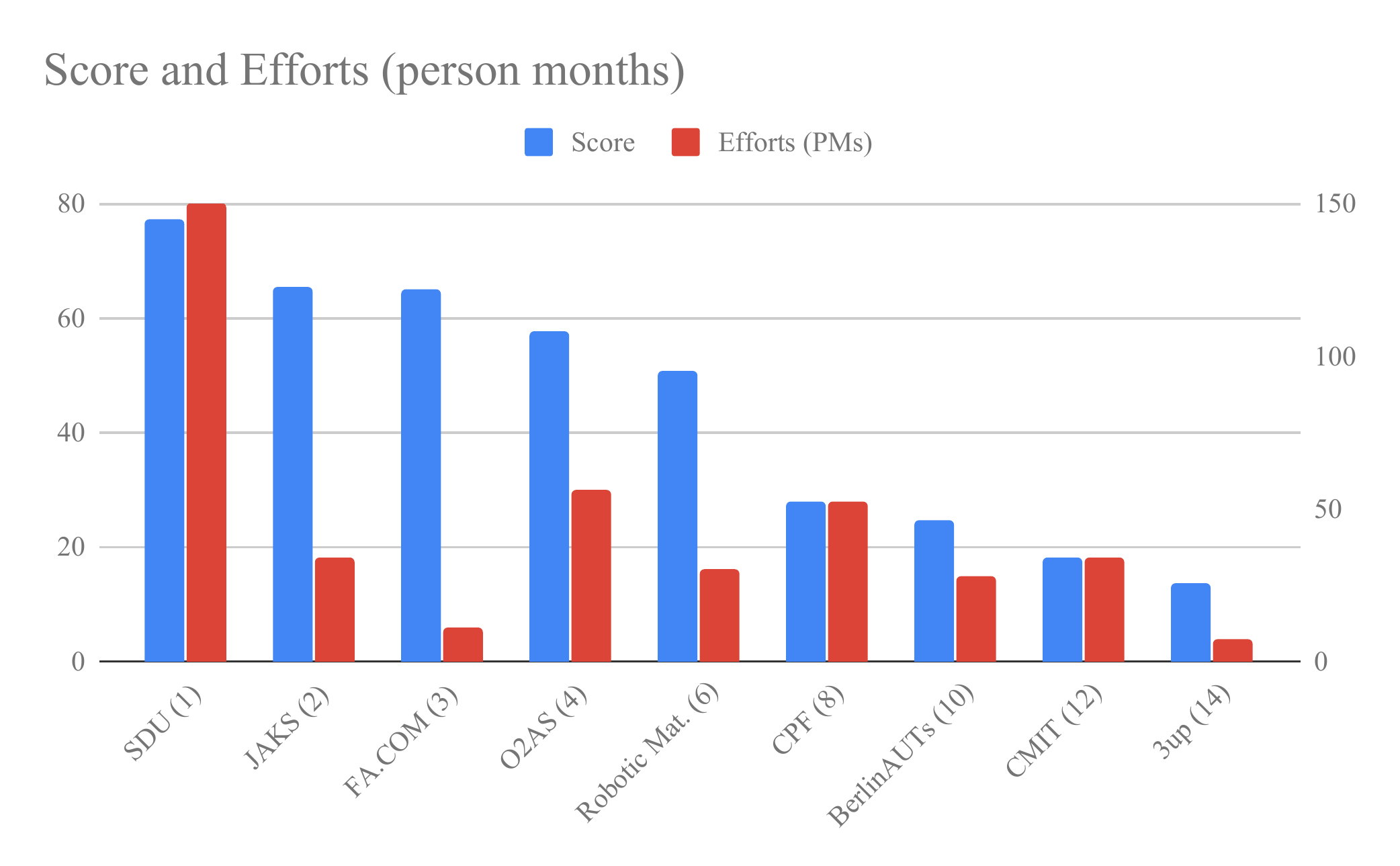}
    \caption{
        Estimated efforts in person months, sorted by score achieved by the team.
    }
    \label{fig:survey_pmsvsscore}
\end{figure}

Figure~\ref{fig:survey_pmsvsscore} depicts the estimated development efforts, given in person months (PM), of the responding teams. 
Estimated PMs and resulting score for the teams do not seem to be correlated. 
An obvious outlier is team SDU Robotics with 80~PMs listed as efforts for 18 members, yielding an average of 4.4~PMs spent by each team member (student helpers, PhD students, postdocs, professors and engineers) during the development time from April to October 2018, including efforts of about 1.5 PMs for attending WRC. 
Another obvious outlier is the industry team FA.COM, who estimated their effort with 6~PMs.
As explained in the next section, FA.COM constructed a conventional robotic cell and used mostly off-the-shelf tools, whereas SDU Robotics designed and developed their custom solution from scratch.
However, we want to stress the fact that the specified efforts are retrospective estimations and thus should only be considered indicative.

\subsubsection{System description}

Of the 8 responding teams (3up did not take part in this section of the survey), 5 teams used robots by Universal Robots (mostly UR5), 2 teams used Yaskawa's GP-7 and one team used FANUC's LR Mate 200iD. 
From the limited data of the survey, a tendency may be deduced that the Japanese teams preferred to apply robots from Japanese manufacturers --- probably mostly due to availability for the competition. 
Hence, the URs were mainly used by the international teams, across all institution types, universities as well as companies (including institutions labelled as ``mixed'').
Most teams applied a combination of two manipulators. 
Only the BerlinAUTs system was starting with a single manipulator, and only O2AS and Robot System Integators used 3 manipulators in their setup.

Regarding grippers, all teams made use of two or more grippers in their systems. 
The large majority of teams used parallel grippers, of which three teams used grippers by Robotiq and three teams developed grippers on their own. 
The parallel gripper used by Team Robotic Materials is equipped with integrated computation and sensors for 3D perception, and has since been developed into a commercial product by the startup company of the same name.
Other grippers in the competition were by Weiss Robotics, OnRobot, and CFK.
While most teams used a single set of finger tips for all tasks in the competition and all actions in the tasks, some teams made use of exchangeable finger tips that were changed manually between tasks (BerlinAUTs), or automatically between actions (SDU RoboticsThere's no easy answer to "is it better to be specialized or a generalist", because it depends on your job and what you do, and want to do. 
It is more common to be specialized and manage other specialists in R\&D or design or technical, but you don't need as much technical depth when you sell, organize or manage the product and people.

I would ask what you're really good at, what comes easy to you that you're better at than others, and what you can maintain your interest in. Talking to people and convincing them? Organizing people and large projects? Synthesizing information and making decisions? Managing Figuring out hard technical problems? Writing code? Writing experiment protocols and testing prototypes on your own? Reading norms, standards and technical documentation? You should have a reasonable idea of this by now.

With that and what you learned about possible positions in your industry and adjacent fields, you should be able to decide on a good (or at least possible) career goal, or at least a direction for where you would like to end up. "Let's see how far I get doing X" is perfectly reasonable to start with, too.

\begin{figure}[!htb]
    \centering
    \includegraphics[width=.7\textwidth]{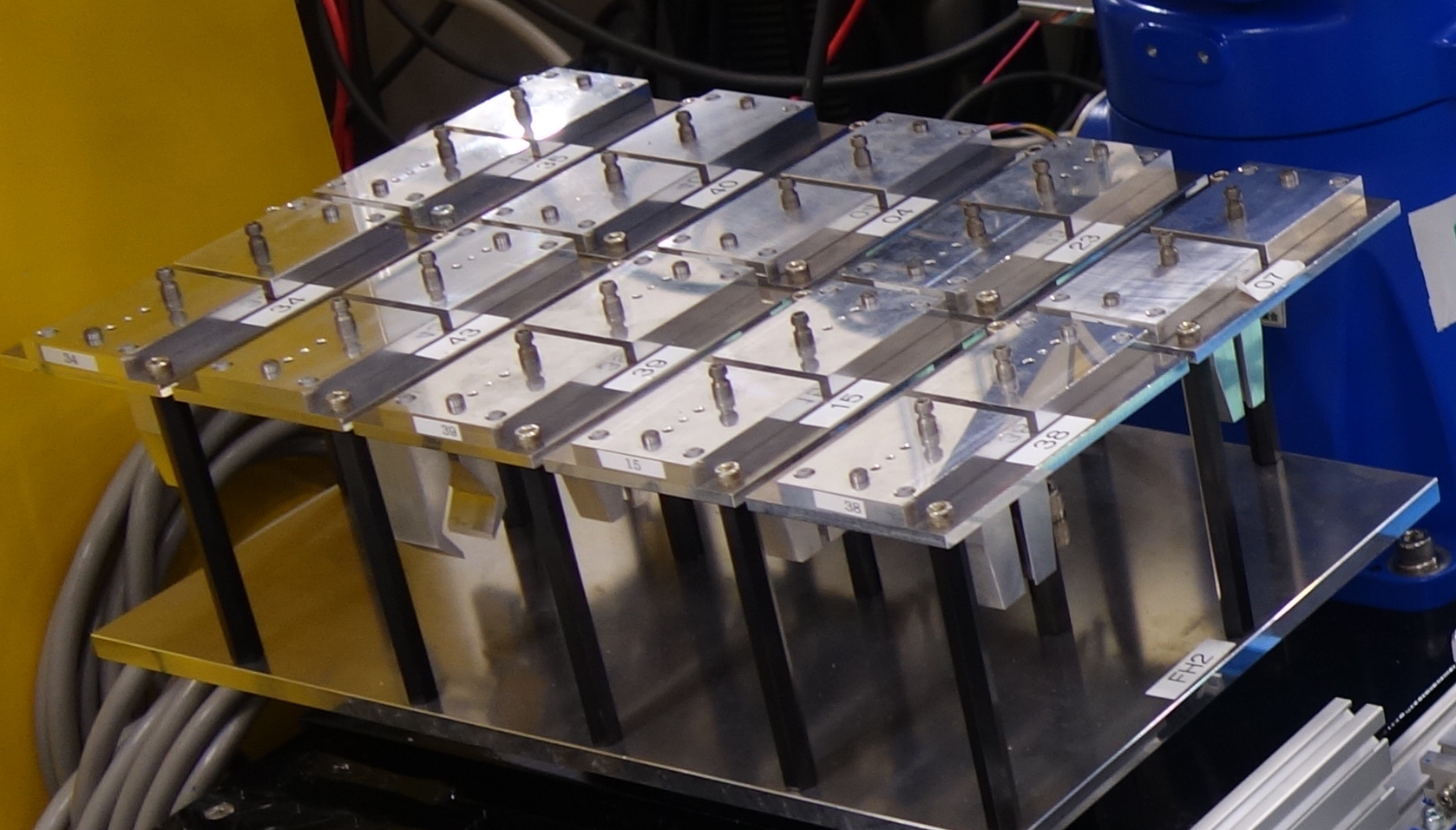}
    \caption{
    A rack of different fingertips used by Team Robot System Integrators.
    }
    \label{fig:fingertip-rack}
\end{figure}

I would be less worried specifically about "being too specialized" with a UX design job than about the general question if it fits your plan of where you want to end up. How will it look in 3-5 years when you want to make your next step? Does it make sense looking back from that point? Will it write a convincing story, or does it sound disjointed and unlikely to give you a good opportunity in another company? It's almost impossible to answer these without knowing where you want to go (or have a few different alternative goals)., Robot System Integrators), as shown in Figure~\ref{fig:fingertip-rack}. 
To cover the range of screw actions in WRC, some teams chose to exchange screw holders and/or screw bits between the screwing actions (e.g. FA.COM and SDU Robotics), while others chose to use differently equipped screwdrivers (e.g. O2AS and CMIT), or grasped screw bits using a centering gripper (e.g. JAKS and Cambridge Robotics).
Instead of making use of conventional industrial screw feeders, JAKS, Robotic Materials and FA.COM developed custom solutions to directly address screw positioning and feeding, too. 

Many teams applied jigs for centering or aligning parts, particularly during the kitting and the assembly tasks.
BerlinAUTs and Robotic Materials used a jig to reorient and position screws: The robot would drop the screw in and then retrieve it in an upright position.
This was particularly relevant for the kitting task, in which the screws must be placed upright in the screw holders of the kitting tray.
In order to facilitate bin picking, SDU Robotics used a system with 3d-printed, part-specific shovels.
The gripper takes some parts on its shovel, then shakes the shovel and uses it similar to a shaker conveyor to retrieve one aligned part from it. The remaining parts could easily be dumped in the bin again.
Further jigs were used in the assembly task, e.g. BerlinAUTs and JAKS used a holder for the motor, as its orientation was important for the assembly with the motor plate. JAKS furthermore used a fixture for the base plate.
Of the 8~responding teams, only O2AS did not use any jigs. 
They made use of their third robot to accomplish alignments of parts, and their custom gripper to align screws.

All teams used 2D~cameras as vision sensors. 3~teams additionally applied sensors with range information (FA.COM, O2AS, Robotic Materials), e.g. time-of-flight or stereo vision cameras by Photoneo and Intel Realsense.
The cameras were mostly used for object detection (SDU Robotics, JAKS, Robotic Materials, BerlinAUTs, O2AS), e.g. to detect the positions of the parts on the placement mat during the taskboard task.
Vision sensors were further used to check if a grasp operation was successful (O2AS), or for more general marker detection (SDU Robotics, BerlinAUTs), e.g. to identify the positioning of the taskboard or kitting trays in a flexible setup.
Some teams did not mention for what operations they used their vision sensors (FA.COM, CPF, CMIT).
Team BerlinAUTs was the only team that performed visual servoing during peg-in-hole assembly operations. 
However, they were also the only team that did not apply a force sensor.

SDU Robotics, Robotic Materials and CMIT used force-torque sensors for the object insertion tasks.
Another typical application of force sensors was to detect contact, in the manner of moving the end effector downwards until a contact to the taskboard or to the parts in a bin has been detected (O2AS, Robotic Materials).
While SDU Robotics used the integrated force-torque sensors of the UR10e robot, other teams had additional sensors at the wrist, e.g. Optoforce sensor (Robotic Materials), Robotiq FT-300 (CPF, O2AS), or Yaskawa Motofit (CMIT).
Further sensors were used for checking the success of a grasping operation, e.g. a tactile sensor by JAKS, or an air flow sensor in the SDU Robotics vacuum gripper.
O2AS used air pressure sensors in their suction tool, as well as in screw tools with integrated suction.
After screwing, they used internal motor sensors to check if the screw is fastened.




The remaining questions concern the used software framework, motion planner and simulation tools.
Half of the responding teams relied at least partially on ROS (SDU Robotics, JAKS, O2AS, CPF).
Some also added further tools such as Choreonoid (JAKS), a robotics simulator developed by the National Institute of Advanced Industrial Science and Technology (AIST) Japan, 
RobWork (SDU Robotics), an open-source collection of C++ libraries for simulation and control of robot systems developed by SDU Robotics, and VEROSIM (SDU), a 3d~simulation software.
Two teams used a service-oriented, distributed architecture based on XML RPC (Robotic Materials) or gRPC (BerlinAUTs).
CMIT stayed in line with their use of Yaskawa products and relied on the MotoCom SDK.
Custom motion planners were used by SDU Robotics, JAKS, CPF and CMIT. 
O2AS used MoveIt, a motion planning framework for ROS. 
Robotic Materials and BerlinAUTs used only the Cartesian path inverse kinematics provided by the Universal Robot.

Simulating the behavior of robotic systems with high complexity and dependency on sensor input can be a daunting task itself.
Most teams applied offline simulations (SDU Robotics, CMIT, JAKS, CPF, O2AS) with the help of the aforementioned software frameworks. 
O2AS additionally performed online simulation, and SDU could simulate crucial components in-the-loop.
BerlinAUTs only used the URSim module to simulate robot movement, but without any sensor input. 
FA.COM and Robotic Materials did not use any simulation tools.

\subsubsection{Competition Conditions}
Table~\ref{tab:efforts-for-subtasks} illustrates how much effort the teams spent on each individual subtask.
We can observe that most efforts were spent to solve the 2D~assembly (taskboard) task, which was the task where teams also earned most points.
Particularly the lower ranked teams concentrated on the taskboard, presumably because it was the easiest task to begin with.
Only FA.COM focused most on the assembly task and consequently they were the only team to successfully assemble one belt drive unit to completion during the competition.
However, some teams argued that the most significant effort was developing the base system itself rather than programming single tasks.
Having a flexible system setup allows to solve a range of tasks with less task-specific programming or adaptations.
This is reflected by the answers to the question of what changes were made to the system setup in between tasks, as most teams applied, if any, only relatively small changes such as exchanging finger tips, tools or grippers.

\begin{table}[htbp]
  \centering
  \caption{Distribution of efforts on the individual subtasks. Percentages are reported. (Note that the percentages of CPF have been recomputed as the original values did not add up to 100\%.)}
    \begin{tabular}{|l|r|r|r|r|r|}
    \hline
    Team    & Rank   & Taskboard & Kitting & Assembly & Surprise \\
    \hline
    SDU Robotics        & 1 & 40    & 30    & 25    & 5 \\
    JAKS                & 2 & 30    & 30    & 30    & 10 \\
    FA.COM              & 3 & 40    & 10    & 50    & 0 \\
    O2AS                & 4 & 40    & 20    & 40    & 0 \\
    Robotic Materials   & 6 & 40    & 40    & 15    & 5 \\
    CPF                 & 8 & 50    & 14.3  & 21.4  & 14.3 \\
    BerlinAUTs          & 10 & 75    & 20    & 5     & 0 \\
    CMIT                & 12 & 75    & 10    & 10    & 5 \\
    \hline
    Average             &    & 48.8 & 21.8 & 24.6 & 4.9 \\
    \hline
    \end{tabular}%
  \label{tab:efforts-for-subtasks}%
\end{table}%

Regarding the strategies for the hardest task, the surprise assembly aiming towards ``level 5'' automation, many teams simply dealt with it on the day of the competition.
The parts' specifications were given to the competitors one day before, and the actual parts were handed out 2~hours prior to their trial.
This made it possible for many teams to teach some movements and perform at least some of the required assembly operations.
Only three of the teams that participated in the survey had more sophisticated strategies.
Robotic Materials had primitives to grasp unknown objects and expected those to transfer directly to the surprise parts.
Furthermore, they used simple assembly primitives written in Python, which they could quickly adapt.
O2AS updated the models of the assembled parts, which updated their software with the new target positions.
They aimed for mostly using the same code as in the assembly, just with updated target positions.
Team SDU Robotics relied on a reusable, modular framework to reprogram or adapt workcells, which separates between hardware (exchangeable 3D printed fingers and fixtures), software (reusable SkillBlocks representing the component functionalities) and communication (unified component calls via ROS ActionServers / OPC Unified Architecture).
The gripper-finger exchange system in combination with the 3D~printers they brought to the competition allowed them to create custom fingers for the new parts.
Using play-back methods for quick programming of movements and agile pose estimation algorithms based on CAD models helped to implement assembly operations with surprise parts.

The overall strategies of the teams are particularly interesting.
FA.COM for example did not aim for points, but heavily focused on successfully finishing the assembly.
By contrast, Robotic Materials focused on the task board and kitting tasks to learn general primitives for assembly and bin picking.
They expected the primitives to transfer to the assembly task, but ended up running out of preparation time. 
SDU Robotics, JAKS, Cambridge, and O2AS instead have prepared equally for all tasks. 

Many teams identified compliance as a critical element of their technical strategy to make up for perception uncertainty, which is in contrast to rigging all parts as precisely and accurate as possible, such as in FA.COM and Robotic System Integrators' approaches, which could be labeled as ``conventional robotic workcells''. 
Specifically Robotic Materials, JAKS, O2AS, BerlinAUTs, and CMIT each identify using some combination of vision, feedback control, and compliance to create reusable software components that are robust to uncertainty in part location. 
SDU Robotics chose a middle ground, by relying both on rigging and feedback control and introduced flexibility by 3D printing appropriate fingers and tools, facilitated by their digital-twin programming approach. 
None of the teams relied exclusively on robotic grippers, but either used tool changers or used their grippers to pick up additional tools such as screwdrivers (Robotic Materials, O2AS, e.g.) or suction cups (O2AS).

Regarding the system components that caused the most trouble during the competition, the teams reported a wide variety of challenges. 
Teams that heavily relied on rigging reported long downtimes in between switching from kitting to assembly tasks (SDU Robotics).
Teams relying on a more autonomous approach based on feedback control had trouble with inaccuracies introduced by compliance (JAKS), camera calibration (O2AS) and lighting conditions (CPF), and calibration of the robotic arms (O2AS, CMIT). 
Teams also reported systemic challenges such as breaking hardware (BerlinAUTs), perception and control algorithms not being adept at the tasks (BerlinAUTs, Robotic Materials), or collaborative robotic arms not being able to provide enough contact forces during insertion tasks due to safety features or insufficient control loop frequency (O2AS). 

\subsubsection{Takeaways}

Finally, we asked the teams to relay the most important lesson learned during the competition.
While some teams note that certain components need to be improved in particular, such as a feedback controller (JAKS) or gripper design (CPF), many other teams realized that we are still far away from solutions reaching the robustness and repeatability required for industrial use (O2AS, Robotic Materials, SDU Robotics).
Particularly the robustness and autonomous error handling and recovery were mentioned as future challenges by O2AS and BerlinAUTs.
SDU Robotics even proposed to use the same tasks again for a subsequent competition for measuring progress.
BerlinAUTs and FA.COM emphasized that the large variety of different approaches pursued by all the teams was impressive.
O2AS point out that there is no established standard way to solve the task, that things that seem to be simple sometimes turn out to be very hard, and that the effort required to set up the system, connect the modules and program the tasks seems excessive, considering the level of sophistication of the underlying hardware.

\subsection{Analysis}

Although a single competition cannot be a reliable indicator of a specific technology's performance, we nevertheless observe the following trends. 
Competitors have been in two camps: those constructing conventional robotic work cells using more or less exclusively off-the-shelf tools (FA.COM, Robotic System Integrators, e.g.) and those who mostly rely on custom-made end-effectors and novel software approaches (JAKS, O2AS, and Robotic Materials, e.g.). 
It might not be a coincidence that the winning team SDU Robotics chose a middle-ground and combined top-of-the-line industrial collaborative robotic arms, 3D perception systems, tool changers, and fixtures with custom designed digital twin software that allowed them to generate 3D-printed fixtures and assembly code with just one day turn-around time. 

With both camps performing reasonably well, one might argue that the jury is still out on whether a conventional approach or an ``autonomous'' one that relies on heavy sensing and feedback control, will ultimately lead to more versatile manufacturing systems. 
We need to note, however, that we are comparing mature industrial products, which have been in development for multiple decades, with a new class of systems, which has been developed completely from scratch (JAKS, O2AS, Robotic Materials, Cambridge), except for the industrial arms themselves. 

The design of the challenge may also have contributed to the fact that both camps could perform reasonably well.
Whereas the kitting task and the surprise assembly are clearly targeted towards ``level~5'' automation, the other tasks were defined beforehand and did not introduce major uncertainties during the trials.
This was further stressed by the fact that all teams had two trials, in between which the teams could use all provided parts for further optimization, including teaching positions, which suited most of the conventional robotic cells.
On the other hand, the kitting and surprise assembly tasks were difficult for the more autonomous assembly systems as well.
We suggest that putting more emphasis on the difficult tasks would foster solutions with a higher level of autonomy.

\section{Discussion}
There was only one system (FA.COM), which performed the complete 3D assembly. 
Albeit this approach lacked in versatility with regards to kitting and 2D assembly, conventional robotic work cells remain still dominant when it comes to automated assembly of large quantities of an item. 
Adding autonomy added to such systems, for example by relaxing position constraints on parts, adding sensing and feedback control, will surely reduce setup time, but may even end up reducing cycle times, which become the dominant metric as quantities increase. 

With consumers demanding more and more goods and new products at a faster pace, and customization increasing, we expect industry to increase the amount of autonomy in assembly robots. 
Particularly promising here is SDU Robotics' approach that seamlessly integrates robot programming and generating 3D-printed fixtures from part 3D models and vision co-located with the robotic grippers as employed by many teams (SDU Robotics, JAKS, O2AS, Robotic Materials, e.g.).
Mounting the camera on the gripper reduces the calibration problem vs. an externally mounted camera, and the resulting information can not only be used for feedback control, but also for error detection.

It is noteworthy that the cost estimates of the systems do not include the decades of R\&D efforts that have gone into the development of current-generation robotic systems. 
All of the systems with a higher degree of autonomy were individual solutions that required new research and development efforts and were developed from scratch, which resulted in a higher cost in terms of both money and time.
The fact that no established standard approaches or frameworks exist for flexible, autonomous assembly systems exacerbated this effect.
FA.COM, which used readily available technology, is one of the teams which spent the least time developing their systems. 
On the other hand, SDU Robotics, which developed their custom-made system from scratch, spent roughly almost ten times more PMs compared to FA.COM.

In addition, we observed only few common points between the more autonomous solutions, e.g. the widespread usage of ROS and/or collaborative robots such as UR5.
This could be explained by the need for rapid prototyping over maximum performance, the desire to minimize implementation work, the need for a support community, and for a "standard" solution with low barriers to entry.
There is an evident unsatisfied desire in the community for a "standard" software approach to robotics, but evaluating the many available frameworks and middlewares is out of the scope of this paper.
Team SDU Robotics expressed the ambition to develop a platform not only for the World Robot Competition~2018 but also for present and future research and development projects with industrial partners, to speed up the implementation of new tasks.

On the topics of task representations or languages for assembly and manipulation, as well as multi-step assembly planning, we note that while there are some lines of research aiming at the topic, none of the teams implemented such an approach in the competition.
It is unclear if this implies a) that these methods are not mature or useful yet, b) that the task was so small that no abstraction or automated planning was necessary, or c) that constructing the systems took all the teams' available workload.

Only one team (BerlinAUTs) explicitly mentions ``machine learning'' and only in the context of object detection. 
Siemens AG has posited a similar assembly problem as machine learning challenge\footnote{\url{https://www.siemens.com/us/en/home/company/fairs-events/robot-learning.html}, accessed 2019-07-02.}. 
As in the WRS, the task contains a variety of round and square peg-in-hole problems, but is further complicated by multiple interlocking gears. 
These kind of tasks put further emphasis on force control, which can be used to detect the correct insertion point. 
Although it is an interesting challenge for so-called ``end-to-end learning'', i. e. learning task completion without any prior information, the industrial framing of the WRS and the fact that standardized and well-known parts are concerned does not suggest the need for machine learning at first sight, but rather the use of robust, rule-based algorithms or assembly primitives, as these methods depend less on time-intensive optimization and parameter tuning.
Rather than end-to-end learning these algorithms, it is more likely that complex assembly behaviors could be learned by exploring the discrete space of existing primitives. 
Here, automatically generating appropriate assembly sequences from a CAD model of the assembly is probably most promising in the context of an assembly challenge such as at WRS.

Assembly is a contact-rich task which requires precise object localization, which means that it cannot be solved by ``soft'' manipulation techniques like suction cups or caging grippers.
The assembly challenge thus complements logistics and household-themed pick-and-place tasks, which in the past have often become perception challenges, as the amount of object manipulation is so minimal that it can usually be performed with suction cups and the like. 
In current industrial assembly robots, the need for precise localization is commonly solved by custom made jigs that position each part.
As planning with jigs and well-defined part positions is already well understood, future challenges could stress versatility and level 5 automation, for example by requiring teams to assembly different objects from similar parts, or the belt drive from a variety of interchangeable parts (which appears to be the trend expressed in the announcement of the WRC 2020).

Furthermore, from a product designer's perspective, it might be valuable to explore the degrees of freedom of the aforementioned assembly primitives.
For example, a difference in radius of pulleys may be more easily adapted to than switching from a rubber belt to metal chain.
While on the one hand, automatic manufacturing technologies should be made more agile, on the other hand, pushing the design towards exploiting cheaper changes which are in line with potential assembly primitives could facilitate the manufacturing of these products.

Despite assembly being a contact-rich task, force and tactile sensing was almost unused beyond wrist- or joint-based impedance control, contact sensing, or setting the grasp force of grippers.
Only JAKS used a tactile sensor to detect grasp success and to avoid grasping multiple belts.
This is somewhat surprising, given the importance of tactile feedback and control in humans.
It is our interpretation that tactile sensors are not sufficiently mature for use in industrial applications, and suspect that a lack of durability, resolution and/or ease of purchase is an obstacle to a more widespread use.

\section{Conclusion}

The assembly competition at the World Robot Summit 2018 has challenged the community to create a robotic solution for the autonomous assembly, which required screwing actions, the insertion of machine parts with realistic tolerances, and the tensioning of a rubber belt.
The tasks were of an appropriate difficulty, so that only one team finished a complete assembly, with many coming close.
Yet, the most successful 3D assembly solution has been relying on commercially available off-the-shelf robots and tools, calling into question the competition format's ability to push for transformative departures from the state of the art.
We conclude that:
\begin{itemize}
    \item The set-up costs for new systems are still very high, especially on the software side.
    \item Autonomy and new algorithms did not correlate with success in the assembly task
    \item The overall degree of autonomy was still low (little error recovery, almost no multi-stage planning)
    \item New approaches, such as full 3D simulation of the assembly, hand-held tools and eye-in-hand and self-centering grippers are promising
    \item Tactile sensing on fingertips made no real appearance, only within wrists and joints.
\end{itemize}

In terms of notable or promising approaches, we found:
\begin{itemize}
    \item Modeling the complete assembly using 3D simulation tools (``digital twin''), to allow on-the-fly adaptation and motion generation
    \item Tools held/grasped by the robot gripper and 3D printed, part-specific tools
    \item Gripper-mounted 3D cameras, which reduce the calibration error compared to scene-mounted cameras, and allow the picking of very small objects
\end{itemize}
Although the competition in its current form was sufficient to showcase and possibly disseminate such technology, a format that emphasizes versatility, for example by focusing on constructing a variety of similar items, might be better suited to promote jump innovations. 


Felix von Drigalski is a senior researcher at OMRON SINIC X Corporation in Tokyo. He obtained MSc-level engineering diplomas from the Karlsruhe Institute of Technology in Germany and the INSA Lyon in France, and a PhD in Computer Science at the Nara Institute of Science and Technology in Japan. His research interests include robotic manipulation, tactile sensing and automated assembly. His teams obtained 1st place at the Airbus Shopfloor Challenge at ICRA 2016, Finalist at the Amazon Robotics Challenge 2017, and 4th place and the Special Award of the SICE at the World Robot Summit Assembly Challenge 2018.

Nikolaus Correll is an Associate Professor of Computer Science, at the University of Colorado Boulder. He obtained a degree in electrical engineering from ETH Zürich in 2003, a PhD in Computer Science from EPFL in 2007, and did a post-doc at MIT’s Computer Science and Artificial Intelligence lab from 2007-2009. He is the recipient of a 2012 NSF CAREER award, a 2012 NASA Early Career fellowship, and a Provost’s Achievement award.

Martin Rudorfer graduated in computational engineering science at Technical University Berlin in 2014. From 2014 to 2019 he has been research and teaching assistant at the department for industrial automation technology of TU Berlin, where he is also a PhD candidate. His main research interests are object detection and pose estimation for robotic manipulation.

Christian Schlette is a Professor at the Maersk Mc-Kinney Moeller Institute (MMMI) at the University of Southern Denmark (SDU).
In 2006, Christian joined the foundation of Institute for Man-Machine Interaction (MMI) at RWTH Aachen University, Germany as principal investigator and received his doctorate summa cum laude (Dr.-Ing.) from RWTH Aachen University in 2012. 
He joined SDU Robotics in 2017.
In his research, he is interested in synergies across the fields of automation, robotics, simulation and control in application areas such as factory automation and large structure production.

Joshua C. Triyonoputro is a graduate student at the Graduate School of Engineering Science, Osaka University. 
He joined the Robotic Manipulation Laboratory in 2018 to research industrial automation. 
His main research interests are lean assembly automation and non-prehensile manipulation using robotic arms.

Tokuo Tsuji received his BS, MS, and Doctoral degrees from Kyushu University in 2000, 2002 and 2005, respectively. 
He worked as a Research Fellow of Graduate School of Engineering, Hiroshima University from 2005 to 2008. 
He worked as a Research Fellow of Intelligent Systems Research Institute of National Institute of Advanced Industrial Science and Technology (AIST) from 2008 to 2011. 
From 2011 to 2016, he worked as a Research Associate at Kyushu University. 
From 2016, he has been working as an Associate Professor at Institute of Science and Engineering, Kanazawa University. 
His research interest includes multi-fingered hands, machine vision, and software platforms of robotic systems. 
He is a member of IEEE, JSME, RSJ, and IEICE.

Weiwei Wan received a Ph.D. degree from the Department of Mechano-Informatics, Graduate School of Information Science and Engineering, University of Tokyo, Tokyo, Japan, in 2013.
From 2013 to 2015, he was a Post-Doctoral Research Fellow of the Japanese Society for the Promotion of Science, Japan, and did his Post-Doctoral study in the Robotics Institute, Carnegie Mellon University, Pittsburgh, PA, USA. 
He was a Research Scientist with the National Institute of Advanced Industrial Science and Technology, Tsukuba, from 2015 to 2017, and is now an associate professor at the Graduate School of Engineering Science, Osaka University. 
His research interests include robotic hands, robotic manipulators, and robotic grasping and manipulation planning for next-generation manufacturing.

Tetsuyou Watanabe received the B.S., M.S., and Dr. Eng. degrees in mechanical engineering from Kyoto University, Kyoto, Japan, in 1997, 1999, and 2003, respectively. 
From 2003 to 2007, he was a Research Associate with the Department of mechanical Engineering, Yamaguchi University, Japan. 
From 2007 to 2011, he was an assistant professor with Division of Human and Mechanical Science and Engineering, Kanazawa University. 
From 2011 to 2018, he was an associate professor with Faculty of Mechanical Engineering, Institute of Science and Engineering, Kanazawa University. 
Since 2018, he has been a professor with Faculty of Frontier Engineering, Institute of Science and Engineering, Kanazawa University. 
From 2008 to 2009, he was a visiting researcher at Munich University of Technology. 
His current research interests include mechanics and control of robotic systems, medical sensors, surgical systems, grasping, and object manipulation. 
He is a member of the IEEE, JSME, SICE, RSJ, and JSMBE.


\bibliography{ma}

\label{lastpage}

\end{document}